\documentclass[10pt,journal,compsoc]{IEEEtran}

\ifCLASSOPTIONcompsoc
  \usepackage[nocompress]{cite}
\else
  \usepackage{cite}
\fi
\ifCLASSINFOpdf
\else
\fi

\usepackage{epsfig}
\usepackage{graphicx}
\usepackage{amsmath}
\usepackage{amssymb}
\usepackage{booktabs}
\usepackage{amsfonts}
\usepackage{float}
\usepackage[caption = false, font=footnotesize]{subfig}
\usepackage{array}
\usepackage{color}
\usepackage{tabularx}
\usepackage{longtable}
\usepackage{tabu}
\usepackage{multirow}
\usepackage{url}
\definecolor{grey}{RGB}{160,160,160}   
\definecolor{black}{RGB}{0,0,0} 
\definecolor{red2}{RGB}{255,0,0} 
\definecolor{red}{RGB}{0,0,0} 
\usepackage{bm}
\usepackage[noend]{algpseudocode}
\usepackage{algorithmicx,algorithm}

\makeatletter
\newcommand*{\rom}[1]{\expandafter\@slowromancap\romannumeral #1@}
\makeatother

\newcolumntype{L}[1]{>{\raggedright\let\newline\\\arraybackslash\hspace{0pt}}m{#1}}
\newcolumntype{C}[1]{>{\centering\let\newline\\\arraybackslash\hspace{0pt}}m{#1}}
\newcolumntype{R}[1]{>{\raggedleft\let\newline\\\arraybackslash\hspace{0pt}}m{#1}}

\newcommand{\etal}{\textit{et al}.}
\newcommand{\ie}{\textit{i}.\textit{e}.}
\newcommand{\eg}{\textit{e}.\textit{g}.}

\newcommand{\blockc}[3]{\multirow{3}{*}{\(\left[\begin{array}{c}\text{3$\times$3, #1, stride #2}\\[-.1em] \text{3$\times$3, #1, stride #2} \end{array}\right]\)$\times$#3}
}

\newcommand{\blockd}[5]{\multirow{3}{*}{\(\left[\begin{array}{c}\text{3$\times$3, #2, stride #3}\\[-.1em] \text{3$\times$3, #2, stride #4} \end{array}\right]\)$\times$#5}
}

\begin{document}
%
\title{Continual Learning for Blind Image \\ Quality Assessment}

\author{Weixia~Zhang,~\IEEEmembership{Member,~IEEE,}
        Dingquan~Li, 
        Chao~Ma,~\IEEEmembership{Member,~IEEE,}
        Guangtao~Zhai,~\IEEEmembership{Senior Member,~IEEE,}
        Xiaokang~Yang,~\IEEEmembership{Fellow,~IEEE,}
        and~Kede~Ma,~\IEEEmembership{Member,~IEEE}
\IEEEcompsocitemizethanks{\IEEEcompsocthanksitem W. Zhang, C. Ma, G. Zhai, and X. Yang are with the MoE Key Lab of Artificial Intelligence, AI Institute, Shanghai Jiao Tong University, China.\protect\\
E-mail: {\{zwx8981, chaoma, zhaiguangtao, xkyang}\}@sjtu.edu.cn.

\IEEEcompsocthanksitem D. Li is with Peng Cheng Laboratory, Shenzhen, China.\protect\\
E-mail: lidq01@pcl.ac.cn.

\IEEEcompsocthanksitem K. Ma is with the Department of Computer Science, City University of Hong Kong, Kowloon, Hong Kong.\protect\\
E-mail: kede.ma@cityu.edu.hk.}
\thanks{Corresponding author: Kede Ma.}
}


\IEEEtitleabstractindextext{%
\begin{abstract}
The explosive growth of image data facilitates the fast development of image processing and computer vision methods for emerging visual applications, meanwhile introducing novel distortions to processed images. This poses a grand challenge to existing blind image quality assessment (BIQA) models, which are weak at adapting to subpopulation shift. Recent work suggests training BIQA methods on the combination of all available human-rated IQA datasets. However, this type of approach is not scalable to a large number of datasets and is cumbersome to incorporate a newly created dataset as well. In this paper, we formulate continual learning for BIQA, where a model learns continually from a stream of IQA datasets, building on what was learned from previously seen data. We first identify five desiderata in the 
continual setting with three criteria to quantify the prediction accuracy, plasticity, and stability, respectively. We then propose a simple yet effective continual learning method for BIQA. Specifically, based on a shared backbone network, we add a prediction head for a new dataset and enforce a regularizer to allow all prediction heads to evolve with new data while being resistant to catastrophic forgetting of old data. We compute the overall quality score by a weighted summation of predictions from all heads. Extensive experiments demonstrate the promise of the proposed continual learning method in comparison to standard training techniques for BIQA, with and without experience replay. We made the code publicly available at \url{https://github.com/zwx8981/BIQA_CL}.
\end{abstract}

}

\maketitle

\IEEEdisplaynontitleabstractindextext

%
\IEEEpeerreviewmaketitle

\ifCLASSOPTIONcompsoc
\IEEEraisesectionheading{\section{Introduction}\label{sec:introduction}}
\else
\section{Introduction}\label{sec:introduction}
\fi

\begin{figure}[!t]
  \centering
  \includegraphics[width=\columnwidth]{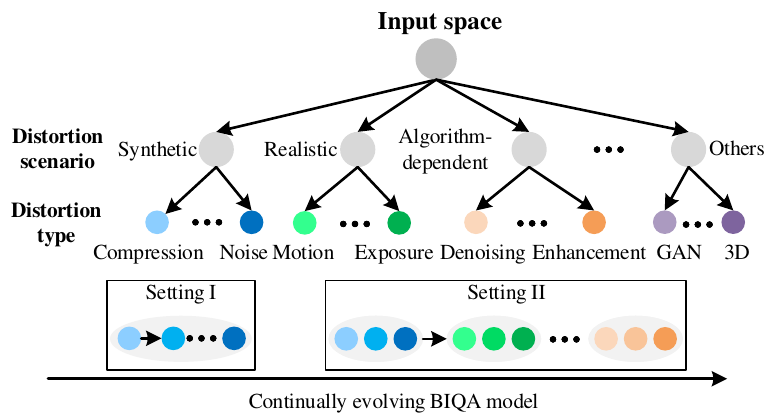}
  \caption{Illustration of the continual learning paradigm for BIQA. Subpopulation shift exists across distortion types and scenarios. In Setting \rom{1}, a BIQA model continually evolves from one distortion type to another within the same  distortion scenario. In  Setting \rom{2}, a BIQA model continually evolves with varying distortion scenarios.}\label{fig:illustration}
\end{figure}

\IEEEPARstart{A}{iming} to automatically quantify human perception of image quality, blind image quality assessment (BIQA)~\cite{wang2006modern} has experienced an impressive series of successes due in part to the creation of human-rated image quality datasets over the years. For example, the LIVE dataset~\cite{sheikh2006statistical} marks the switch from distortion-specific~\cite{wang2002no} to general-purpose BIQA~\cite{mittal2012no,ye2012unsupervised}. The CSIQ dataset~\cite{larson2010most} enables cross-dataset comparison. The TID2013 dataset~\cite{ponomarenko2013color} and its successor KADID-10K~\cite{lin2019kadid} expose the difficulty of BIQA methods in generalizing to different distortion types. The Waterloo Exploration Database~\cite{ma2017waterloo} tests model robustness to diverse content variations of natural scenes. The LIVE Challenge Database~\cite{ghadiyaram2016massive} probes the synthetic-to-real generalization, which is further evaluated by the KonIQ-10K~\cite{hosu2020koniq} and SPAQ~\cite{fang2020perceptual} datasets. Assuming that the input domain $\mathcal{X}$ of BIQA is the space of all possible images, each IQA dataset inevitably represents a tiny \textit{subpopulation} of $\mathcal{X}$ (see Fig.~\ref{fig:illustration}). That is, BIQA models are bound to encounter subpopulation shift when deployed in the real world. It is thus of enormous value to build robust BIQA models to subpopulation shift.

Previous work~\cite{mittal2012no,ye2012unsupervised,bosse2016deep,Ma2018End} on BIQA mainly focuses on boosting performance within subpopulations, while few efforts have been dedicated to testing and improving model robustness to subpopulation shift. Mittal~\etal~\cite{mittal2013making} aimed ambitiously for \textit{universal} BIQA by measuring a probabilistic distance between patches extracted from natural undistorted images and those from the test ``distorted'' image. The resulting NIQE only works for a limited set of distortions. Zhang~\etal~\cite{zhang2015feature} modified NIQE by adding more expressive statistical features with marginal improvement. 

A straightforward adaptation to subpopulation shift is to fine-tune model parameters with new data, which has been extensively practiced by the BIQA methods based on deep neural networks (DNNs). However, new learning may destroy performance on old data, a phenomenon known as \textit{catastrophic forgetting}~\cite{mccloskey1989catastrophic}. Recently, Zhang~\etal~\cite{zhang2020learning,zhang2021uncertainty} proposed a dataset combination trick for training BIQA models against catastrophic forgetting. Despite demonstrated robustness to subpopulation shift, this type of method may suffer from three limitations. First, it is not scalable to handle a large number of datasets because of the computation and storage constraints. Second, it is inconvenient to accommodate a new dataset since training samples from all datasets are required for joint fine-tuning. Third,  some datasets may not be accessible after a period of time (\eg, due to privacy issues~\cite{aljundi2019continual}), preventing na\"{i}ve dataset combination.

\begin{table*}[t]
  \centering
  \caption{Summary of IQA datasets used in our experiments. CLIVE stands for the LIVE Challenge Database. SS: Single stimulus.  DS: Double stimulus. MS: Multiple stimulus. CQR: Continuous quality rating. ACR: Absolute category rating. CS: Crowdsourcing}\label{tab:database}
  \begin{tabular}{lccccccc}
      \toprule
        {Dataset} & \# of Images & \# of Training Pairs & \# of Test Images & Scenario & \# of Types & Testing Methodology& Year \\
     \midrule
        LIVE~\cite{sheikh2006statistical} & 779 & 7,000 & 163 & Synthetic & 5 & SS-CQR &2006 \\
        CSIQ~\cite{larson2010most} & 866 & 8,000 & 173 & Synthetic & 6 & MS-CQR & 2010\\
        BID~\cite{ciancio2011no} & 586 & 10,000 & 117 & Realistic & N.A. & SS-CQR & 2011\\
        CLIVE~\cite{ghadiyaram2016massive} & 1,162 & 20,000 & 232 & Realistic & N.A. & SS-CQR-CS  & 2016 \\
        KonIQ-10K~\cite{hosu2020koniq} & 10,073 & 95,000 & 2,015 & Realistic & N.A. & SS-ACR-CS & 2018\\
        KADID-10K~\cite{lin2019kadid} & 10,125 & 95,000 & 2,000 & Synthetic & 25 & DS-ACR-CS & 2019\\
     \bottomrule
  \end{tabular}
\end{table*}

In this paper, we take steps towards assessing and improving the robustness of BIQA models to subpopulation shift in a \textit{continual learning} setting. The basic idea is that a BIQA model learns continually from a stream of IQA datasets, integrating new knowledge from the current dataset (\ie, plasticity) while preventing the forgetting of acquired knowledge from previously seen datasets (\ie, stability). To make continual learning for BIQA feasible, nontrivial, and practical, we identify five desiderata: 1) common perceptual scale, 2)
robust to subpopulation shift, 3) limited direct access to previous data,  4) no test-time oracle, and 5) bounded memory footprint. Furthermore, we describe a simple yet effective continual learning method for robust BIQA to subpopulation shift. Specifically, based on a shared and continually-updated backbone network, we add a quality prediction head for each new dataset as a way of promoting plasticity for learning new knowledge. Consolidation of previous knowledge is implemented by stabilizing predictions of previous heads. We summarize the current training dataset using $K$-means clustering in feature space and use the learned centroids to compute weights for final quality prediction. 

In summary, our main contributions are threefold.
\vspace{-.2cm}
\begin{itemize}\setlength{\itemsep}{2pt}
    \item We establish the continual learning paradigm for BIQA, in which model robustness to subpopulation shift can be evaluated more directly and practically.
    \item We propose a computational method for continually learning BIQA models, which significantly outperforms standard training techniques for BIQA, with and without experience replay \cite{rebuffi2017icarl}.
    \item We conduct extensive experiments to test various aspects of the proposed method, including plasticity, stability, accuracy, and order-robustness.
\end{itemize}

\section{Related Work}\label{sec:related_work}
In this section, we give an overview of representative IQA datasets as different subpopulations from the image space $\mathcal{X}$ (see Table~\ref{tab:database}). We then discuss the progress of BIQA driven by the construction of IQA datasets. Finally, we review continual learning in a broader context. 

\subsection{IQA Datasets}\label{subsec:iqa_database}
Hamid~\etal~\cite{sheikh2006statistical} conducted the first ``large-scale'' subjective user study of perceptual image quality. The resulting LIVE dataset~\cite{sheikh2006statistical} includes 779 distorted images with five synthetic distortion types at five to eight levels. Single stimulus continuous quality rating (SS-CQR) was adopted to collect the mean opinion scores (MOSs). In 2010, Larson~\etal~\cite{larson2010most}  released the CSIQ dataset, covering 866 images with six synthetic distortions at three to five levels, among which four types are shared by LIVE. A form of the multiple stimulus method was used for subjective testing, where a set of images were linearly displaced according to their perceived quality. The horizontal distance between every pair of images reflected the perceptual difference. In 2011, Ciancio~\etal~\cite{ciancio2011no} built the BID dataset, including
mostly blurry images due to camera and/or object motion during acquisition. The same subjective method as in LIVE was adopted to acquire human quality annotations. In 2013, Ponomarenko \etal~\cite{ponomarenko2013color} extended the TID2008 dataset to TID2013 with 3,000 images distorted by 25 types at five levels. Paired comparison with a Swiss-system tournament was implemented to reduce subjective cost. In 2016, Ghadiyaram and Bovik~\cite{ghadiyaram2016massive} created the LIVE Challenge Database with 1,162 images, undergoing complex realistic distortions. They designed an online crowdsourcing system to gather MOSs using the SS-CQR method. In 2017, Ma~\etal~\cite{ma2017waterloo} complied the Waterloo Exploration Database, aiming to probe model generalization to image content variations.
No subjective testing was conducted. Instead, the authors proposed three rational tests, namely, the pristine/distorted image discriminability test (D-Test), the listwise ranking consistency test (L-Test), and the pairwise preference consistency test (P-Test) to evaluate IQA methods more economically. From 2018 to 2019, two large-scale datasets, KADID-10K~\cite{lin2019kadid} and KonIQ-10K~\cite{hosu2020koniq}, were made publicly available, which significantly expand the number of synthetically and realistically distorted images, respectively. MOSs of the two datasets were sourced on crowdsourcing platforms using a single stimulus absolute category rating.
In 2020, Fang~\etal~\cite{fang2020perceptual} constructed the SPAQ dataset for perceptual quality assessment of smartphone photography. Apart from MOSs, EXIF data, image attributes, and scene category labels were also recorded to facilitate the development of BIQA models for real-world applications. Concurrently, Ying~\etal~\cite{ying2020from} built a large dataset that contains patch quality annotations.

As discussed previously, different datasets may use different subjective procedures, leading to different perceptual scales of the collected MOSs. Even if two datasets happen to use the same subjective method, their MOSs may not be comparable due to differences in the purposes of the studies and the visual stimuli of interest. In Sections~\ref{sec:cl_biqa}~and~\ref{sec:new_algorithm}, we will give a careful treatment of this subtlety in continual learning for BIQA.

\subsection{BIQA Models}\label{subsec:biqa_methods}
In the pre-dataset era, the research in BIQA dealt with specific distortion types, such as JPEG compression~\cite{wang2002no} and JPEG2000 compression~\cite{marziliano2004perceptual}. Since the inception of the LIVE dataset, general-purpose BIQA began to be popular. Many early methods relied on natural scene statistics (NSS) extracted from either spatial domain~\cite{mittal2012no,mittal2013making} or transform domain~\cite{moorthy2011blind,saad2012blind}. The underlying assumption is that a measure of the destruction of statistical regularities of natural images~\cite{simoncelli2001natural} provides a reasonable approximation to perceived visual quality. Another line of work explored unsupervised feature learning for BIQA~\cite{ye2012unsupervised,xu2016blind}. Since the introduction of the LIVE Challenge Database, synthetic-to-real generalization of BIQA models has received much attention. Ghadiyaram and Bovik \cite{ghadiyaram2017perceptual} handcrafted a bag of statistical features specifically for authentic camera distortions. As the number of images in the newly released IQA datasets became larger, deep learning came into play and began to dominate the field of BIQA. 
Many strategies were proposed to compensate for the lack of human-labeled data, including patchwise training~\cite{kang2014convolutional, bosse2016deep}, transfer learning~\cite{zeng2018blind}, and quality-aware pre-training~\cite{liu2017rankiqa, Ma2018End, zhang2020blind,ma2019blind, 9121773}. To confront the synthetic-to-real challenge (and vice versa), Zhang~\etal~\cite{zhang2020learning, zhang2021uncertainty} proposed a computational method of training BIQA models on multiple datasets. Latest interesting BIQA studies include active learning for improved generalizability~\cite{wang2020active}, meta-learning for fast adaptation~\cite{zhu2020metaiqa}, patch-to-picture mapping for local quality prediction~\cite{ying2020from}, loss normalization for accelerated convergence~\cite{li2020norm}, and adaptive convolution for content-aware quality prediction~\cite{9156687}.

\subsection{Continual Learning}\label{subsec:cl}
Human learning is a complex and incremental process that continues throughout the life span.
While humans may forget the learned knowledge, they forget it gradually rather than catastrophically~\cite{french1999catastrophic}. However, this is not the case for machine learning models such as DNNs, which tend to completely forget old concepts once new learning starts~\cite{mccloskey1989catastrophic}. A plethora of continual learning methods has been proposed, mainly in the field of image classification.
Li and Hoiem~\cite{li2017learning} proposed learning without forgetting (LwF), which uses model predictions of previous tasks as pseudo labels in a knowledge distillation framework~\cite{hinton2015distilling}.
Based on LwF, Rannon~\etal~\cite{rannen2017encoder} attempted to alleviate domain shift among tasks in the learned latent space. Aljundi~\etal~\cite{aljundi2017expert} introduced a set of gating autoencoders for automatically feeding a sample to the relevant expert network during inference. Another family of methods identifies and penalizes changes to important parameters with respect to previous tasks when learning new tasks.
Representative work includes elastic weight consolidation~\cite{kirkpatrick2017overcoming} and its online variant~\cite{schwarz2018progress}, incremental moment matching~\cite{lee2017overcoming}, variational continual learning~\cite{nguyen2018variational}, synaptic intelligence~\cite{zenke2017continual}, and memory-aware synapses~\cite{aljundi2018memory}. Masse~\etal~\cite{masse2018alleviating} proposed context-dependent gating as a complementary module to  weight consolidation ~\cite{kirkpatrick2017overcoming, zenke2017continual}. 


With the increasing length of task sequence, soft regularization techniques may not suffice
to constrain the model parameters in feasible regions. A plausible solution is parameter isolation~\cite{delange2021continual} as a form of hard regularization, which allows growing branches to accommodate new tasks~\cite{rusu2016progressive} or masking learned parameters for previous tasks~\cite{fernando2017pathnet, mallya2018packnet, mallya2018piggyback}.
While parameter isolation effectively prevents catastrophic forgetting, it requires the task oracle to activate the corresponding branch or mask during inference.

Experience replay methods are data-level continual learning solutions, which store old samples or generate pseudo-samples with generative models~\cite{delange2021continual}. As a simple baseline, experience replay~\cite{rolnick2019experience} has been combined with the reservoir sampling~\cite{vitter1985random}. Rebuffi~\etal~\cite{rebuffi2017icarl} developed a class incremental learner, iCaRL, which stores a subset of exemplars per class for representation replay. During inference, iCaRL calculates the mean of each class in the learned feature space, and performs the nearest-mean-of-exemplars classification. Lopez-Paz~\etal~\cite{lopez2017gradient} proposed gradient episodic memory (GEM), which was improved by Chaudhry~\etal~\cite{chaudhry2019efficient} in terms of efficiency. Aljundi~\cite{aljundi2019online} proposed to replay samples
whose predictions will be most negatively impacted by the foreseen parameter updates (\ie, the most interfered). Most recently, Prabhu~\etal~\cite{prabhu2020gdumb} pointed out the caveats in the progress of continual learning for classification. They proposed a na\"{i}ve method that greedily stores samples in the memory buffer, based on which a model is trained to achieve state-of-the-art performance.

It is important to note that the recent success of continual learning for image classification may not transfer in a straightforward way to BIQA. This motivates us to establish a continual learning paradigm for BIQA, identifying desiderata to make it feasible, nontrivial, and practical. We also contribute to effective and robust continual learning methods for training BIQA models.

\section{A Continual Learning Paradigm for BIQA}\label{sec:cl_biqa}
In this section, we formulate continual learning for BIQA with five desiderata and three evaluation criteria to quantify the prediction accuracy, plasticity, and stability, respectively.

\begin{figure*}[!t]
    \centering
    \subfloat[$\mathrm{MOS} = 25$]{\includegraphics[width=0.249\textwidth]{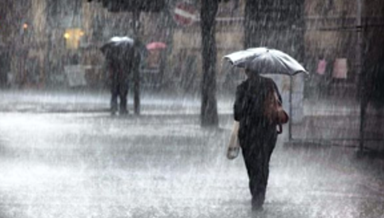}}\hfill
      \subfloat[$\mathrm{MOS} = 39$]{\includegraphics[width=0.249\textwidth]{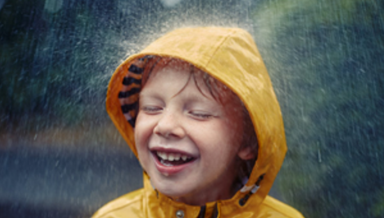}}\hfill
    \subfloat[$\mathrm{MOS} = 51$]{\includegraphics[width=0.249\textwidth]{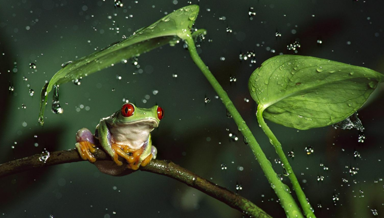}}\hfill
    \subfloat[$\mathrm{MOS} = 69$]{\includegraphics[width=0.249\textwidth]{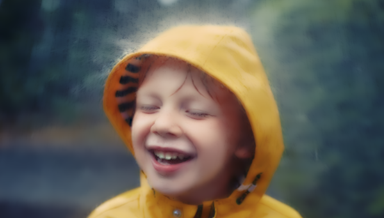}}\hfill
    \caption{Images sampled from the deraining quality assessment dataset~\cite{wu2020subjective}. A larger MOS in the dataset denotes lighter rain density. It is not hard to observe that rain density is not monotonically correlated with perceived image quality. Therefore, the dataset violates Desideratum \rom{1} and should be excluded from the task sequence for BIQA. Images are cropped for improved visibility.}\label{fig:derain}
\end{figure*}

\subsection{Problem Definition}~\label{subsec:new_setting}
We define the learning on a new IQA dataset as a new task in our continual learning setting. When training on the $t$-th dataset $\mathcal{D}_t$, we assume no direct access to $\{\mathcal{D}_k\}_{k=1}^{t-1}$ for the moment, leading to the following training objective:
\begin{align}\label{eq:tra}
\mathcal{L}(\mathcal{D}_t;w) =\frac{1}{\vert\mathcal{D}_t\vert}\sum_{(x,q)\in \mathcal{D}_t}\ell(g_w(x),q),
\end{align}
where $x\in\mathbb{R}^N$ and $q\in\mathbb{R}$ denote the $N$-dimensional ``distorted'' image  and the corresponding MOS, respectively. $g_w$ represents a computational BIQA model parameterized by a vector $w$.
$\ell(\cdot)$ is the objective function, quantifying the quality prediction performance. One may add a regularizer $r(w)$ to Eq.~\eqref{eq:tra} with the goal of gaining resistance to catastrophic forgetting. For evaluation, we may measure the performance of $g_w$ on the hold-out test sets of all tasks seen so far: 
\begin{align}\label{eq:def}
\sum_{k=1}^{t}\mathcal{L}(\mathcal{V}_k;w) = \sum_{k=1}^{t}\left(\frac{1}{\vert\mathcal{V}_k\vert}\sum_{(x,q)\in \mathcal{V}_k}\ell(g_w(x),q) \right),
\end{align}
where $\mathcal{V}_k$ is the test set for the $k$-th task.
An ideal BIQA model should perform well on new tasks, and endeavor to mitigate catastrophic forgetting of old tasks, resulting in a low objective value in Eq.~\eqref{eq:def}.

\subsection{Five Desiderata}~\label{subsec:desiderata}
Considering the differences between image classification and BIQA, we argue that careful treatment should be given to make continual learning for BIQA feasible, nontrivial, and practical.
Towards this, we list five desiderata.

\begin{itemize}
    \item[\textbf{\rom{1}}] \textbf{Common Perceptual Scale}. This requires that the MOSs of IQA datasets should admit a common perceptual scale. In other words, there exists a \textit{monotonic} function for each dataset to map its MOSs to this common scale. Otherwise, learning a single $g_w$ on multiple datasets continually is conceptually infeasible. Desideratum \rom{1} excludes human-rated datasets that measure perceptual quantities closely related to image quality (\eg, visual contrast~\cite{wang2008maximum} and scene visibility~\cite{choi2015referenceless, wu2020subjective}). To highlight this point, we show some images from the deraining quality assessment dataset~\cite{wu2020subjective} in Fig.~\ref{fig:derain}, with a  smaller MOS indicating severer rain density. It is clear that rain density is not \textit{monotonically} correlated with visual quality. Therefore, this dataset violates Desideratum \rom{1}, and should be excluded from the task sequence for BIQA.\\
   \item[\textbf{\rom{2}}]  \textbf{Robust to Subpopulation Shift}. It is empirically proven that existing BIQA models generalize reasonably to distortions with similar visual appearances (\eg, from additive noise to multiplicative noise; from LIVE~\cite{sheikh2006statistical} to CSIQ~\cite{larson2010most}). However, when datasets exhibit apparent subpopulation shift (\eg, synthetic to realistic distortions), the generalization of BIQA models remains particularly weak. Although the stream of training data may be distorted in arbitrary form (\eg, with one distortion type only), it is highly desirable to develop continual learning methods that are robust to various levels of (and especially apparent) subpopulation shift.\\
   \item[\textbf{\rom{3}}] \textbf{Limited Access to Previous Data}. This is the key desideratum~\cite{delange2021continual} that makes continual learning continual learning. Experience replay continual learning relies on replaying (at least a small portion of) old data to fight against catastrophic forgetting~\cite{rebuffi2017icarl,lopez2017gradient, chaudhry2019efficient,hayes2020remind,pellegrini2020latent, hou2019learning,Buzzega2020dark}. In the context of BIQA, Zhang~\etal~\cite{zhang2020learning,zhang2021uncertainty} proposed to jointly train models on data from all tasks, which can be seen as a performance upper bound~\cite{delange2021continual}. Desideratum \rom{3} assumes limited access to previous data when training new tasks, meaning that the memory buffer should be carefully controlled within a preset budget. It puts no constraints on the format of old data, which can either be raw images in previous datasets or their feature summaries.\\
   \item[\textbf{\rom{4}}] \textbf{No Test-Time Oracle}. As advocated in~\cite{delange2021continual,farquhar2018towards}, a well-designed continual learning method should be independent of the task oracle to make predictions. That is, the method should be unaware of which dataset the test image belongs to. Desideratum \rom{4} is imperative in BIQA because if we know in advance the task label, we may be able to train separate and specialized models for each of the datasets, making continual learning for BIQA a trivial task.\\
  \item[\textbf{\rom{5}}] \textbf{Bounded Memory Footprint}. The model capacity in the number of model parameters should be relatively fixed, forcing the BIQA method to allocate its capacity wisely to achieve the Pareto optimum between plasticity and stability. Other memory overhead, \eg, to store old data and/or pseudo-labels should also remain bounded, or at least grow very slowly, with respect to the number of tasks seen so far~\cite{rebuffi2017icarl}.
\end{itemize}

\begin{figure*}
  \centering
  \includegraphics[width=0.95\textwidth]{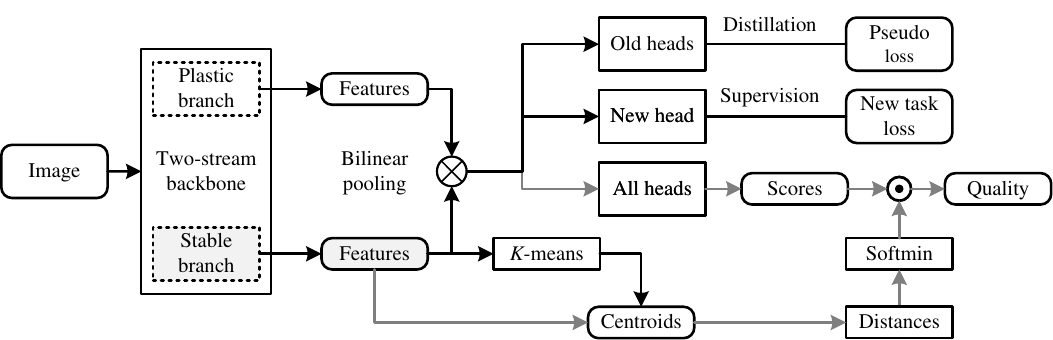}
  \caption{System diagram of the proposed continual learning method. Black and grey arrows correspond to the training and testing phases, respectively.}\label{fig:lfque}
\end{figure*}

\subsection{Performance Measures}~\label{subsec:metric}
We propose three quantitative criteria to measure 1) the prediction accuracy, 2) the plasticity, and 3) the stability of a BIQA
model in the continual learning setting. Conceptually, plasticity and stability refer to the ability to integrate new information and preserve previous knowledge, respectively. 
Without loss of generality, we use Spearman's rank correlation coefficient (SRCC) between model predictions and MOSs as a measure of prediction monotonicity. Other correlation measures (\eg, Kendall rank correlation coefficient and Pearson linear correlation coefficient) and distance metrics (\eg, mean squared error and mean absolute error) can also be applied.

For a BIQA model trained on a sequence of $T$ tasks (denoted by the $T$-th model), we compute the mean SRCC between model predictions and MOSs of each dataset as a measure of prediction accuracy:
\begin{equation}\label{eq:msrcc}
    \mathrm{mSRCC}  = \frac{1}{T}\sum_{k=1}^{T}\mathrm{SRCC}_{Tk},
\end{equation}
where $\mathrm{SRCC}_{tk}$ is the SRCC result of the $t$-th model on the $k$-th dataset. We then define a mean plasticity index ($\mathrm{mPI}$):
\begin{equation}\label{eq:plasticity}
    \mathrm{mPI}  =\frac{1}{T}\sum_{t=1}^{T}\mathrm{PI}_{t}= \frac{1}{T}\sum_{t=1}^{T}\mathrm{SRCC}_{tt},
\end{equation}
\ie, the average result of the model on the current dataset along the task sequence. Last, we define a mean stability index ($\mathrm{mSI}$) by measuring the variability of model performance on old data when new learning has taken place:
\begin{equation}\label{eq:stability1}
\mathrm{mSI} = \frac{1}{T}\sum_{t=1}^{T}\mathrm{SI}_{t},
\end{equation}
where
\begin{equation}\label{eq:stability2}
\mathrm{SI}_{t} =\begin{cases}
    1 &  t=1 \\
    \frac{1}{t-1}\sum_{k=1}^{t-1}\widehat{\mathrm{SRCC}}_{tk} & t > 1 
  \end{cases},
\end{equation}
where $\widehat{\mathrm{SRCC}}_{tk}$ for $k < t$ is computed between the predictions of the $t$-th and $k$-th models. It is noteworthy that a higher $\mathrm{mSI}$ does not necessarily imply better quality prediction performance on old tasks as no MOS is involved. $\mathrm{mSRCC}$, $\mathrm{mPI}$, and $\mathrm{mSI}$ measure different and complementary aspects of a continually learned BIQA model. We may further define a mean plasticity-stability index ($\mathrm{mPSI}$) over a list of $T$ tasks to quantify the plasticity-stability trade-off:
\begin{equation}\label{eq:psi}
\mathrm{mPSI} =\frac{1}{T}\sum_{t=1}^{T}\mathrm{PSI}_{t} =\frac{1}{2T}\sum_{t=1}^{T}\left(\mathrm{PI}_{t} + \mathrm{SI}_{t}\right).
\end{equation}

\section{A Continual Learning Method for BIQA}~\label{sec:new_algorithm}
In this section, we propose a simple yet effective continual learning method for BIQA. The system diagram of our method is shown in Fig.~\ref{fig:lfque}. 

\subsection{Model Estimation}\label{subsec:specification}
We describe the proposed method with respect to the desiderata stated in Subsection~\ref{subsec:desiderata}. 
\subsubsection{Learning-to-Rank for BIQA}\label{subsubsec:d1}
According to Desideratum \rom{1}, it is desirable to work with a common perceptual scale. However, this is difficult because we are only given a stream of $T$ IQA datasets without the monotonic functions to map the associated MOSs. Inspired by~\cite{zhang2020learning,zhang2021uncertainty}, we choose to learn a perceptual scale for all tasks by exploiting relative quality information. Specifically, given an image pair $(x, y)$,  we compute a binary label:
\begin{align}\label{eq:bgt}
     p(x,y) = 
\begin{cases} 
 1 & \mbox{if } q(x)\ge q(y) \\
      0 & \mbox{otherwise} \end{cases},
\end{align}
where $q(x), q(y)\in\mathbb{R}$ are the MOSs of images $x$ and $y$, respectively. When learning the $t$-th task, we transform $\mathcal{D}_t =\{x^{(i)}_t, q^{(i)}_t\}_{i=1}^{\vert\mathcal{D}_t\vert}$ to $\mathcal{P}_t=\{(x^{(i)}_t,y^{(i)}_t), p^{(i)}_t\}_{i=1}^{M_t}$, where $M_t \le \binom{\vert\mathcal{D}_t\vert}{2}$.

Our BIQA model consists of a backbone network, $f_\phi: \mathbb{R}^N \mapsto \mathbb{R}^L$, that takes the $N$-dimensional raw image as input and produces an $L$-dimensional feature vector. We append a prediction head, $h_{\psi_t}(\cdot)$ parameterized by $\psi_{t}$, to compute quality estimates for the $t$-th task. Under the Thurstone's Case V model, we are able to estimate the probability that $x$ is of higher quality than $y$ by
\begin{align}\label{eq:difference2}
\hat{p}_t(x,y)= \Phi\left(\frac{h_{\psi_t}(f_{\phi}(x)) - h_{\psi_{t}}(f_{\phi}(y))}{\sqrt{2}}\right),
\end{align}
where $\Phi(\cdot)$ is the standard Normal cumulative distribution function, and the variance of quality predictions is fixed to one \cite{thurstone1927law}. The full set of parameters, $\{\phi,\psi_1,\psi_2,\ldots, \psi_T\}$, constitute the parameter vector $w$ to be optimized.

For the current $t$-th task, we measure the statistical distance between the ground-truth and predicted probabilities using the fidelity loss~\cite{tsai2007frank}, whose advantages over the cross-entropy loss have been demonstrated in several BIQA studies~\cite{zhang2020learning, zhang2021uncertainty}:

\begin{align}\label{eq:fidelity}
\ell_{\mathrm{new}}(x, y;\phi, \psi_t)
= 1& - \sqrt{p(x,y)\hat{p}_t(x,y)} \nonumber \\ &-\sqrt{(1-p(x,y))(1-\hat{p}_t(x,y))}.
\end{align}

\subsubsection{Mitigating Catastrophic Forgetting}\label{subsubsec:mitigate_cf}
Direct optimization of Eq.~\eqref{eq:fidelity} may cause catastrophic forgetting of old tasks (see Table~\ref{tab:srcc_results}). According to Desideratum \rom{3}, we consider two cases: 1) no previous data is directly accessed, and the model can only train on new data~\cite{delange2021continual}; 2) there is a preset memory budget for storing a small portion of samples from previous tasks. The preserved old data can be used for experience replay (also called rehearsal) to confront the catastrophic forgetting.

We begin with the simpler replay-free case. Inspired by LwF~\cite{li2017learning}, we add a regularizer to allow forgetting old knowledge gracefully. Before training the $t$-th task, we use the $k$-th output head to compute a probability $\bar{p}_{tk}(x,y)$ for each pair of $(x,y)\in \mathcal{P}_t$ according to Eq.~\eqref{eq:difference2}. This creates $t-1$ datasets $\{\mathcal{P}_{tk}\}_{k=1}^{t-1}$ with pseudo-labels to constrain the updated prediction $\hat{p}_{tk}$ to be close to the recorded prediction $\bar{p}_{tk}$. Again, we use the fidelity loss to implement the constraint:
\begin{align}\label{eq:fidelityc}
\ell_{\mathrm{old}}\left(x,y;\phi, \{\psi_k\}_{k=1}^{t-1}\right)& 
=\sum_{k=1}^{t-1}\left( 1- \sqrt{\bar{p}_{tk}(x,y)\hat{p}_{tk}(x,y)}\right. \nonumber \\ &\hspace{-4.5mm}\left.-\sqrt{(1-\bar{p}_{tk}(x,y))(1-\hat{p}_{tk}(x,y))}\right).
\end{align}
In practice, we randomly sample a mini-batch $\mathcal{B}_t$ from $\mathcal{P}_t$ and
use a variant of stochastic gradient descent to minimize the following empirical loss:
\begin{align}\label{eq:minibatch}
\mathcal{L}\left(\mathcal{B}_{t};\phi, \{\psi_k\}_{k=1}^{t}\right) =&\frac{1}{\vert\mathcal{B}_t\vert}\sum_{(x,y)\in\mathcal{B}_t}\left(\ell_{\mathrm{new}}(x, y;\phi, \psi_t)\right. \nonumber\\
&\left.+\lambda\ell_{\mathrm{old}}\left(x,y;\phi, \{\psi_k\}_{k=1}^{t-1}\right)\right),
\end{align}
where $\lambda$ governs the trade-off between the two terms.

There is abundant wisdom in leveraging experience replay to improve continual learning for classification~\cite{mai2021online}. We will adapt and evaluate representative methods for BIQA in Subsection~\ref{subsec:replay}.

\subsubsection{Network Specification}\label{subsubsec:network}
We adopt a two-stream network as the backbone, which is adapted from DB-CNN~\cite{zhang2020blind}, a state-of-the-art BIQA model. Our backbone is composed of two branches, a VGG-like CNN and a variant of ResNet-18, with specifications given in Table~\ref{table:network}. Following the practice in~\cite{zhang2020blind}, we pre-train the VGG-like CNN using a large-scale image set with nine synthetic distortion types at two to five degradation levels. This can be formulated as a multiclass classification problem, training the network to discriminate distortion types and levels. It is empirically shown that the learned  features through this process are distortion-aware.
During continual learning, we fix the VGG-like CNN (denoted as the stable branch in Fig.~\ref{fig:lfque}). In contrast, all parameters of the ResNet-18 variant are learnable, denoted as the plastic branch. The input image is fed into both branches, whose outputs are bilinearly pooled to obtain a fixed-length feature vector~\cite{lin2015bilinear}.
\begin{align}\label{eq:bp}
{f}_\phi(x) = {f_{\phi_{p}}(x)}^{T}f_{\phi_{s}}(x),
\end{align} 
where $\phi_p$ and $\phi_s$ are the parameters of the plastic and stable branches, respectively.
We append an $\ell_2$-normalization layer~\cite{wang2017normface} on top of the backbone network:
\begin{align}\label{eq:l2norm}
\tilde{f}_\phi(x) = \frac{f_\phi(x)}{\|f_\phi(x)\|_2},
\end{align}
to project the feature representation onto the unit hypersphere. This pushes the predictions of all heads to a similar range, making subsequent computation (\eg,  weighted summation of quality scores) more numerically stable.

\subsection{Model Inference}\label{subssec:train_inference} During inference, the original LwF for image classification needs the task oracle, which violates Desideratum \rom{4} and is not applicable to BIQA. Instead of relying on the task oracle to precisely activate a task-specific prediction head, we design a $K$-means gating (KG) mechanism to compute a weighted summation of quality estimates from all heads as the overall quality score. 
\begin{table}[t]
	\centering
	\caption{The proposed two-stream network, consisting of a VGG-like CNN~\cite{zhang2020blind} and a variant of ResNet-18~\cite{he2016deep}, for a $T$-length task sequence. The nonlinear activation, max pooling, and normalization layers are omitted}
	\begin{tabular}{c |c }
		\toprule
		VGG-like CNN & ResNet-18 Variant\\
		\hline
		3$\times$3, 48, stride 1 & {7$\times$7, 64, stride 2} \\
		\hline	
        \multirow{3}{*}{3$\times$3, 48, stride 2}& \blockc{64}{1}{2}\\
        & \\
        &\\
		\hline
        \multirow{6}{*}{\shortstack{3$\times$3, 64, stride 1 \\ \\ 3$\times$3, 64, stride 2}}& \blockd{64}{128}{2}{1}{1}\\
        &\\
        &\\
        &\blockc{128}{1}{1}\\
        &\\
        &\\
        \hline
       \multirow{6}{*}{\shortstack{3$\times$3, 64, stride 1 \\ \\ 3$\times$3, 64, stride 2}}& \blockd{128}{256}{2}{1}{1}\\
        &\\
        &\\
        & \blockc{256}{1}{1}\\
        & \\
        &\\
        \hline
        \multirow{6}{*}{\shortstack{3$\times$3, 128, stride 1 \\ \\ 3$\times$3, 128, stride 1 \\ \\ 3$\times$3, 128, stride 2}}& \blockd{256}{512}{2}{1}{1}\\
        & \\
        &\\
        & \blockc{512}{1}{1}\\
        & \\
        &\\
        \hline
	   \multicolumn{2}{c}{Bilinear Pooling}\\
		 \hline
		 Full Connection &  65,536 $\times T$\\
		\bottomrule
	\end{tabular}
	\label{table:network}
\end{table}

During the $t$-th task learning, we compute the fixed-length quality representations $\{\tilde{f}_{\phi_s}(x_t^{(i)})\}_{i=1}^{\vert\mathcal{D}_t\vert}$ by a feedforward sweep of $\mathcal{D}_t$:
\begin{align}\label{eq:l2norm2}
\tilde{f}_{\phi_s}(x) = \frac{\mathrm{pool}({f}_{\phi_s}(x))}{\|\mathrm{pool}({f}_{\phi_s}(x))\|_2},
\end{align}
where $\mathrm{pool}(\cdot)$ denotes
global average pooling over spatial locations. Similar to Eq.~\eqref{eq:l2norm}, we normalize the pooled representation to make it more comparable across different tasks. We then summarize $\mathcal{D}_t$ with $K$ centroids $\{c_{t}^{(j)}\}_{j=1}^{K}$ by applying $K$-means clustering~\cite{lloyd1982least} to $\{\tilde{f}_{\phi_s}(x^{(i)}_t)\}_{i=1}^{\vert\mathcal{D}_t\vert}$. As the number bits to store $K$ centroids is considerably smaller than storing raw training images, Desideratum \rom{3} is respected. We use $f_{\phi_{s}}$ as the feature extractor to distill $\mathcal{D}_t$ because it is fixed and distortion-aware, which effectively reduces the \textit{task-recency} bias~\cite{masana2020class}.
We measure the perceptual relevance of the test image $x$ to $\mathcal{D}_t$ by computing the minimal Euclidean distance between its feature representation  and the $K$ centroids of $\mathcal{D}_t$:
\begin{align}\label{eq:min_distance}
d_{t}(x) = \min_{1\le j\le K}\|\tilde{f}_{\phi_s}(x) - c_{t}^{(j)}\|_2.
\end{align}
We then pass $\{d_t(x)\}_{t=1}^T$ to a softmin function to compute the weight for the $t$-th prediction head:
\begin{align}\label{eq:softmin}
a_{t}(x) = \frac{{\rm exp}(-{\tau}d_{t}(x))}{\sum_{k=1}^{T}{\rm exp}(-{\tau}d_{k}(x))},
\end{align}
where $\tau \geq 0$ is a temperature parameter used to tune the smoothness of the softmin function. 
The final quality score is defined as the inner product between two vectors of relevance weights and quality predictions:
\begin{align}\label{eq:quality}
\hat{q}(x) = \sum_{t=1}^{T}a_{t}h_{\psi_{t}}(f_\phi(x)).
\end{align}
A final note is that the number of parameters of the $T$ prediction heads grows slowly compared to that of the backbone network\footnote{In our implementation, each new prediction head is implemented by a fully connected layer with 65,536 parameters, accounting for less than $0.6\%$ of the total parameters.}. It is important to note that the proposed KG mechanism is orthogonal to the selection of backbone networks and the number of parameters introduced by a new head can be further optimized based on backbone network selection. Meanwhile, the computations of the backbone network are shared across all prediction heads (see Fig.~\ref{fig:lfque}), which means that the computational overhead introduced by a new prediction head is negligible. Thus, our BIQA model meets Desideratum \rom{5} and is rather scalable in terms of the number of training datasets.

\begin{table}[t]
  \centering
  \caption{Performance comparison in terms of $\mathrm{mSRCC}$, $\mathrm{mPI}$, $\mathrm{mSI}$, and $\mathrm{mPSI}$. All methods are trained in chronological order}\label{tab:mpsr}
  \begin{tabular}{lcccc}
      \toprule
      {Method} & $\mathrm{mSRCC}$ & $\mathrm{mPI}$ & $\mathrm{mSI}$ & $\mathrm{mPSI}$\\
           \hline
     MH-CL-O & 0.7672 & 0.8794 & 0.9243 & 0.9019\\
     EWC-O & \textbf{0.8529} & 0.8754 & 0.9559 & 0.9157\\
     SI-O & 0.8344 & \textbf{0.8809} & 0.9474 & 0.9142\\
     MAS-O & 0.8354 & 0.8741 & 0.9553 & 0.9147\\
     LwF-O & 0.8485 & 0.8765 & \textbf{0.9862} & \textbf{0.9314}\\
    \hline
     SL & 0.6767 & 0.8761 & 0.8045 & 0.8403\\
    
     SH-CL & 0.7091 & 0.8778 & 0.8493 & 0.8636\\
       
     MH-CL & 0.7079 & 0.8794 & 0.8103 & 0.8449\\
     MH-CL-KG & 0.7225 & 0.8694 & 0.9086 & 0.8890\\
          
     EWC & 0.6815 & 0.8754 & 0.8132 & 0.8443\\
     EWC-KG & 0.7919 & 0.8607 & 0.9418 & 0.9013\\
        
     SI & 0.7323 & \textbf{0.8809} & 0.8224 & 0.8517\\
     SI-KG & 0.7848 & 0.8765 & 0.9315 & 0.9002\\
    
     MAS & 0.6170 & 0.8741 & 0.7974 & 0.8358\\
     MAS-KG & 0.7795 & 0.8619 & 0.9433 & 0.9026\\
     
     LwF & 0.6693 & 0.8765 & 0.8331 & 0.8548\\
\hline
     \textbf{Proposed (LwF-KG)} & \textbf{0.8150} & 0.8563 & \textbf{0.9796} & \textbf{0.9180}\\
     \bottomrule
  \end{tabular}
\end{table}

\section{Experiments}\label{sec:exp}
In this section, we describe a realistic and challenging experimental setup for continual learning of BIQA models, which strictly obeys Desiderata \rom{1} and \rom{2}. We divide the experiments into two parts according to the accessibility to previous data. As the proposed continual learning method is the first of its kind, the performance comparison is done mainly with respect to its variants, some of which can be treated as performance upper bounds.

\begin{figure}[t]
  \centering
  \includegraphics[width=\columnwidth]{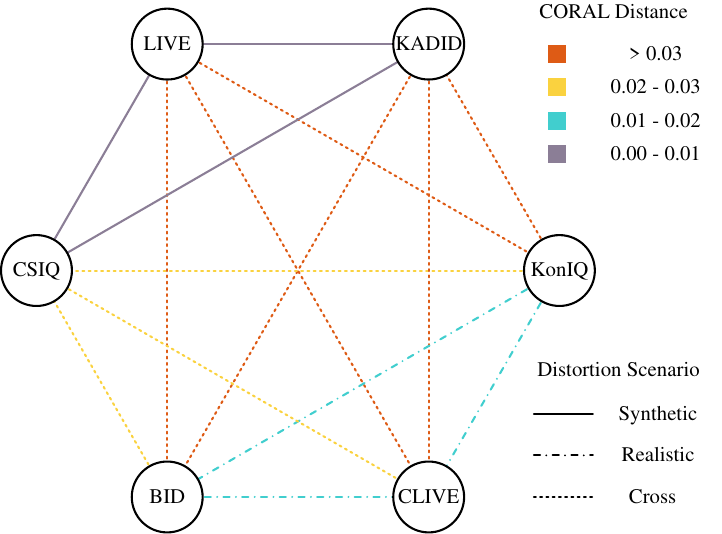}
  \caption{The pairwise CORAL distances between the six IQA datasets with different distortion scenarios. A larger CORAL value indicates a higher level of dissimilarity between two datasets.}\label{fig:coral_graph}
\end{figure}

\subsection{Experimental Setup}~\label{subsec:setup}
We select six widely used IQA datasets, including LIVE~\cite{sheikh2006statistical}, CSIQ~\cite{larson2010most}, BID~\cite{ciancio2011no}, LIVE Challenge~\cite{ghadiyaram2016massive}, KonIQ-10K~\cite{hosu2020koniq}, and KADID-10K~\cite{lin2019kadid}, whose details are summarized in Table~\ref{tab:database}. We leverage the CORrelation ALignment (CORAL)~\cite{sun2016deep}, a widely used non-parametric distance metric to measure the domain similarity, to quantify the dissimilarity between datasets. We define the CORAL loss as the distance between second-order statistics (i.e., covariances) of the VGG-like features of two datasets. As shown in Fig.~\ref{fig:coral_graph}, we find that datasets with synthetic distortions (LIVE, CSIQ, and KADID-10K) are closer to each other compared to those with realistic distortions (and vice versa). We organize these datasets in chronological order, \ie, LIVE $\rightarrow$ CSIQ $\rightarrow$ BID $\rightarrow$ LIVE Challenge $\rightarrow$ KonIQ-10K $\rightarrow$ KADID-10K. In Subsection~\ref{subsec:ablation}, we also test on task sequences of different orders to probe the order-robustness of the proposed method. We randomly sample $70\%$ and $10\%$ images from each dataset for training and validation, leaving the remaining for testing. We follow~\cite{zhang2020learning, zhang2021uncertainty} to form image pairs in $\{\mathcal{P}_t\}_{t=1}^T$, whose numbers are given in Table~\ref{tab:database}. To ensure content independence in  LIVE, CSIQ, and KADID-10K, we divide the training and test sets according to the reference images. Although in the proposed continual learning setting, test sets of future tasks are assumed to be inaccessible, we consider using them for cross-dataset performance evaluation.

For each task, stochastic optimization is carried out by Adam~\cite{kingma2015adam} with $\lambda=10$ in Eq.~\eqref{eq:minibatch}. The parameters of the tailored ResNet-18 and the prediction heads are initialized by the weights pre-trained on ImageNet~\cite{deng2009imagenet} and the He's method~\cite{he2015delving}, respectively.  We set the initial learning rate to $2\times 10^{-4}$ with a decay factor of $10$ for every three epochs, and we train our method for nine epochs.
A warm-up training strategy is used: only the prediction heads are trained in the first three epochs with a mini-batch size of $128$; for the remaining epochs, we fine-tune the entire network with a mini-batch size of $32$. During training, we re-scale and crop the images to $384 \times 384 \times 3$, preserving the aspect ratio.
During testing, the number of centroids used in $K$-means is set to $K=128$ for all tasks. This results in a memory overhead of $64$ kilobytes to store the centroids.
Empirically, we find that the performance is insensitive to the choice of $K$. We tune the temperature $\tau$ in Eq.~\eqref{eq:softmin} on the validation sets and set it to $\tau = 32$. We test images of original size in all experiments.
\begin{table*}[!htbp]
  \scriptsize
  \centering
  \caption{Performance comparison in terms of SRCC between the proposed method and its variants. Best results in each session are highlighted in bold, while results of future tasks are marked in grey}\label{tab:srcc_results}
  \begin{tabular}{cclcccccc}
      \toprule
     {Sequence} & {Dataset} & {Method} & LIVE~\cite{sheikh2006statistical} & CSIQ~\cite{larson2010most} & BID~\cite{ciancio2011no} & CLIVE~\cite{ghadiyaram2016massive} & KonIQ-10K~\cite{hosu2020koniq} & KADID-10K~\cite{lin2019kadid}\\
     \hline
     --  &  All  & JL & 0.9602 & 0.8480 & 0.8796 & 0.8305 & 0.8841 & 0.8801\\
     \hline
     1\textsuperscript{st} &  LIVE & All & \textbf{0.9555} & {\color{grey} 0.7398} & {\color{grey} 0.5872} & {\color{grey} 0.3988} & {\color{grey} 0.6262} & {\color{grey} 0.5248}\\
        \hline
    \multirow{12}{*}{2\textsuperscript{nd}}  &  \multirow{12}{*}{CSIQ} & SL & 0.9257 & 0.8801 & {\color{grey} 0.4337} & {\color{grey} 0.3037} & {\color{grey} 0.5951} & {\color{grey} 0.5613}\\
      &  & SH-CL & \textbf{0.9560} & 0.8783 & {\color{grey} 0.3926} & {\color{grey} 0.2731} & {\color{grey} 0.5482} & {\color{grey} 0.5763}\\
      &   & MH-CL & 0.9450 & 0.8789 & {\color{grey} 0.4141} & {\color{grey} 0.2510} & {\color{grey} 0.5639} & {\color{grey} 0.5460}\\
     &    & MH-CL-KG & 0.9473 & 0.8812 & {\color{grey} 0.4320} & {\color{grey} 0.2729} & {\color{grey} 0.5672} & {\color{grey} 0.5560}\\
      &   & EWC & 0.9385 & 0.8754 & {\color{grey} 0.4705} & {\color{grey} 0.2821} & {\color{grey} 0.5785} & {\color{grey} 0.5578}\\
    &    & EWC-KG & 0.9409 & 0.8773 & {\color{grey} 0.4910} &  {\color{grey} 0.3047} & {\color{grey} 0.5818} & {\color{grey} 0.5658}\\
     &    & SI & 0.9462 & 0.8837 & {\color{grey} 0.4326} & {\color{grey} 0.2680} & {\color{grey} 0.5703} & {\color{grey} 0.5498}\\
     &    & SI-KG & 0.9497 & \textbf{0.8866} & {\color{grey} 0.4615} & {\color{grey} 0.2871} & {\color{grey} 0.5737} & {\color{grey} 0.5589}\\
     &    & MAS & 0.9441 & 0.8808 & {\color{grey} 0.4433} & {\color{grey} 0.2656} & {\color{grey} 0.5704} & {\color{grey} 0.5498}\\
     &    & MAS-KG & 0.9456 & 0.8831 & {\color{grey} 0.4784} & {\color{grey} 0.2880} & {\color{grey} 0.5740} & {\color{grey} 0.5589}\\
     &    & LwF & 0.9491 & 0.8705 & {\color{grey} 0.4668} & {\color{grey} 0.3201} & {\color{grey} 0.5798} & {\color{grey} 0.5551}\\
     &    & Proposed (LwF-KG)  & 0.9499 & 0.8848 & {\color{grey} 0.5100} & {\color{grey} 0.3779} & {\color{grey} 0.5904} & {\color{grey} 0.5471}\\
        \hline
      \multirow{12}{*}{3\textsuperscript{rd}} &  \multirow{12}{*}{BID} & SL & 0.7451 & 0.7847 & 0.8433 & {\color{grey} 0.7180} & {\color{grey} 0.6944} & {\color{grey} 0.5158}\\
      &   & SH-CL & 0.9088 & 0.8068 & 0.8470 & {\color{grey} 0.6165} & {\color{grey} 0.6967} & {\color{grey} 0.5808}\\
    &     & MH-CL & 0.8201 & 0.8186 & 0.8405 & {\color{grey} 0.6609} & {\color{grey} 0.6958} & {\color{grey} 0.5418}\\
     &    & MH-CL-KG & 0.9341 & \textbf{0.9006} & 0.8420 & {\color{grey} 0.5898} & {\color{grey} 0.6687} & {\color{grey} 0.5636}\\
     &    & EWC & 0.8416 & 0.8146 & 0.8455 & {\color{grey} 0.6593} & {\color{grey} 0.7012} & {\color{grey} 0.5442}\\
     &    & EWC-KG & 0.9355 & 0.8996 & 0.8455 & {\color{grey} 0.5685} & {\color{grey} 0.6779} & {\color{grey} 0.5690}\\
      & & SI & 0.8464 & 0.8312 & 0.8435 & {\color{grey} 0.6496} & {\color{grey} 0.7059} & {\color{grey} 0.5523}\\
     &  & SI-KG & 0.9325 & 0.8968 & 0.8429 & {\color{grey} 0.5710} & {\color{grey} 0.6803} & {\color{grey} 0.5665}\\
     &  & MAS & 0.7830 & 0.8205 & 0.8350 & {\color{grey} 0.6625} & {\color{grey} 0.6933} & {\color{grey} 0.5505}\\
    &  & MAS-KG & 0.9287 & 0.9002 & 0.8351 & {\color{grey} 0.5759} & {\color{grey} 0.6712} & {\color{grey} 0.5781}\\
    &  & LwF & 0.8805 & 0.7931 & 0.8585 & {\color{grey} 0.6799} & {\color{grey} 0.6973} & {\color{grey} 0.4705}\\
     &  & Proposed (LwF-KG)  & \textbf{0.9485} & 0.8895 & \textbf{0.8602} & {\color{grey} 0.4691} & {\color{grey} 0.6426} & {\color{grey} 0.5320}\\
        \hline
       \multirow{12}{*}{4\textsuperscript{th}} & \multirow{12}{*}{CLIVE} & SL & 0.7176 & 0.6055 & 0.8262 & 0.8351 & {\color{grey} 0.7573} & {\color{grey} 0.4394}\\
    &  & SH-CL & 0.8352 & 0.6878 & 0.8356 & \textbf{0.8517} & {\color{grey} 0.7758} & {\color{grey} 0.5090}\\
    &  & MH-CL & 0.7169 & 0.6412 & 0.7735 & 0.8443 & {\color{grey} 0.7548} & {\color{grey} 0.5005}\\
     &  & MH-CL-KG& 0.9504 & 0.8714 & 0.7990 & 0.8194 & {\color{grey} 0.7388} & {\color{grey} 0.5305}\\
     &  & EWC & 0.7377 & 0.6818 & 0.7905 & 0.8476 & {\color{grey} 0.7552} & {\color{grey} 0.5079}\\
     &  & EWC-KG & \textbf{0.9527} & \textbf{0.8886} & 0.8261 & 0.8228 & {\color{grey} 0.7394} & {\color{grey} 0.5414}\\
   & & SI & 0.7133 & 0.6344 & 0.7665 & 0.8470 & {\color{grey} 0.7553} & {\color{grey} 0.5000}\\
    &  & SI-KG& 0.9509 & 0.8702 & 0.7946 & 0.8181 & {\color{grey} 0.7387} & {\color{grey} 0.5341}\\
   & & MAS & 0.7450 & 0.6191 & 0.8163 & 0.8464 & {\color{grey} 0.7560} & {\color{grey} 0.5058}\\
   & & MAS-KG& 0.9491 & 0.8587 & 0.8349 & 0.8290 & {\color{grey} 0.7516} & {\color{grey} 0.5379}\\
   & & LwF & 0.7747 & 0.6025 & 0.8331 & 0.8355 & {\color{grey} 0.7460} & {\color{grey} 0.4619}\\
   &  & Proposed (LwF-KG)  & 0.9337 & 0.8805 & \textbf{0.8515} & 0.7999 & {\color{grey} 0.7279} & {\color{grey} 0.5185}\\
        \hline
      \multirow{12}{*}{5\textsuperscript{th}} & \multirow{12}{*}{KonIQ-10K} & SL & 0.6915 & 0.6019 & 0.7434 & 0.7209 & 0.8953 & {\color{grey} 0.5578}\\
  &  & SH-CL & 0.7294 & 0.6554 & 0.7753 & 0.7155 & 0.8948 & {\color{grey} 0.5532}\\
    &   & MH-CL & 0.7146 & 0.6013 & 0.7698 & 0.7256 & 0.8936 & {\color{grey} 0.5471}\\
    &   & MH-CL-KG & 0.9217 & 0.7508 & 0.7720 & 0.7281 & 0.8889 & {\color{grey} 0.4929}\\
     &   & EWC & 0.6647 & 0.5914 & 0.7858 & 0.7040 & 0.8905 & {\color{grey} 0.5489}\\
      &  & EWC-KG & \textbf{0.9534} & 0.8547 & 0.8374 & 0.7691 & 0.8855 & {\color{grey} 0.5231}\\
    &  & SI & 0.6933 & 0.5596 & 0.7927 & 0.7412 & \textbf{0.8960} & {\color{grey} 0.5447}\\
   & & SI-KG & 0.9357 & 0.8078 & 0.8146 & 0.7618 & 0.8917 & {\color{grey} 0.5101}\\
    &   & MAS & 0.6889 & 0.5491 & 0.7760 & 0.7138 & 0.8796 & {\color{grey} 0.5634}\\
   &  & MAS-KG & 0.9498 & \textbf{0.8712} & \textbf{0.8590} & 0.7712 & \textbf{0.8753} & {\color{grey} 0.5643}\\
     &  & LwF & 0.7324 & 0.6447 & 0.7888 & 0.7055 & 0.8866 & {\color{grey} 0.5489}\\
    &   & Proposed (LwF-KG)  & 0.9193 & 0.8621 & 0.8485 & \textbf{0.7851} & 0.8752 & {\color{grey} 0.5510}\\
        \hline
      \multirow{12}{*}{6\textsuperscript{th}} & \multirow{12}{*}{KADID-10K} & SL & 0.8709 & 0.7664 & 0.6554 & 0.3895 & 0.5307 & 0.8473\\
   &  & SH-CL & 0.8577 & 0.7296 & 0.7062 & 0.4548 & 0.6666 & 0.8394 \\
    &   & MH-CL & 0.8618 & 0.7402 & 0.7094 & 0.4216 & 0.6455 & \textbf{0.8634}\\
     &  & MH-CL-KG & 0.8251 & 0.7333 & 0.7481 & 0.5120 & 0.6871 & 0.8292\\
     &  & EWC & 0.8584 & 0.6631 & 0.6518 & 0.4528 & 0.6251 & 0.8380\\
    &   & EWC-KG & 0.8580 & 0.7275 & \textbf{0.8431} & 0.6991 & 0.8463 & 0.7774\\
    &   & SI & 0.8746 & 0.7632 & 0.7418 & 0.4764 & 0.6785 & 0.8594\\
      &  & SI-KG & 0.8575 & 0.7590 & 0.7904 & 0.6588 & 0.8253 & 0.8180\\
    &    & MAS & 0.8106 & 0.6694 & 0.5300 & 0.3521 & 0.4931 & 0.8469\\
   &  & MAS-KG & 0.8395 & 0.7334 & 0.8252 & 0.6673 & 0.8185 & 0.7933\\
    &   & LwF & 0.8548 & 0.7081 & 0.5775 & 0.4379 & 0.5847 & 0.8525\\
    &   & Proposed (LwF-KG) & \textbf{0.8996} & \textbf{0.7893} & 0.8128 & \textbf{0.7742} & \textbf{0.8520} & 0.7622\\
     \bottomrule
   \end{tabular}
\end{table*}

\subsection{Competing Methods}~\label{subsec:training_methods}
 \textbf{Separate Learning (SL)} is the \textit{de facto} method in BIQA. We train the model with a single prediction head on one of the six training sets by optimizing Eq.~\eqref{eq:minibatch} with $\lambda=0$. Although SL is not a continual learning method, we incorporate it as a reference.\\ 
     \textbf{Joint Learning (JL)} is a recently proposed method~\cite{zhang2020learning, zhang2021uncertainty} to overcome the cross-dataset challenge (as a specific form of subpopulation shift) in BIQA. We train the same model with a single head on the combination of all six training sets by optimizing Eq.~\eqref{eq:minibatch} with $\lambda=0$. With full access to all training data, JL serves as the upper bound of all continual learning methods.\\
     \textbf{Single-Head Continual Learning (SH-CL)} is a baseline of the proposed continual learning method, where the same model with a single head is successively trained on $\{\mathcal{P}_t\}_{t=1}^6$ by optimizing Eq.~\eqref{eq:minibatch} with $\lambda=0$. The difference between SL and SH-CL lies in training the model from scratch for the current task and fine-tuning the model with the initialization provided by the previous task.\\
    \textbf{Multi-Head Continual Learning (MH-CL)} is a multi-head extension of SH-CL. MH-CL adds a prediction head for a new task, and optimizes it for  Eq.~\eqref{eq:minibatch} with $\lambda=0$. It remains to specify the head for final quality prediction. To encourage adaptation to a constantly changing environment, we simply use the latest head to make prediction. MH-CL serves as the baseline for all regularization-based continual learning methods. Meanwhile, we may incorporate the proposed KG mechanism during inference, giving rise to \textbf{MH-CL-KG}. Moreover, we leverage the task oracle to precisely activate the corresponding head for prediction, denoted by \textbf{MH-CL-O}, which may give the performance upper bound for multi-head architectures.\\ 
    \textbf{Learning without Forgetting (LwF)} in BIQA builds upon MH-CL by optimizing Eq.~\eqref{eq:minibatch} with $\lambda=10$. In other words, LwF introduces a stability regularizer to preserve the performance of previously seen data. Same as MH-CL, LwF relies on the latest head for quality prediction.\\
    \textbf{The proposed method (LwF-KG)} can be seen as the combination of LwF and KG. We also explore the task oracle to select the corresponding head for quality prediction, denoted by \textbf{LwF-O}.\\
    \textbf{Parameter Importance Regularization} follows a similar paradigm that penalizes the changes to the estimated "important" parameters for previous tasks when learning a new task. Specifically, we implement three such regularizers - elastic weight consolidation (EWC)~\cite{kirkpatrick2017overcoming}, synaptic intelligence (SI)~\cite{zenke2017continual}, and memory aware synapses (MAS)~\cite{aljundi2018memory}. 
    Similar to LwF, all three methods are built upon the MH-CL baseline with the KG mechanism, denoted by \textbf{EWC-KG}, \textbf{SI-KG}, and \textbf{MAS-KG}, respectively. The task oracle can also be leveraged for quality prediction, denoted by \textbf{EWC-O}, \textbf{SI-O}, and \textbf{MAS-O}, respectively.

\begin{figure}[t]
  \centering
  \includegraphics[width=\columnwidth]{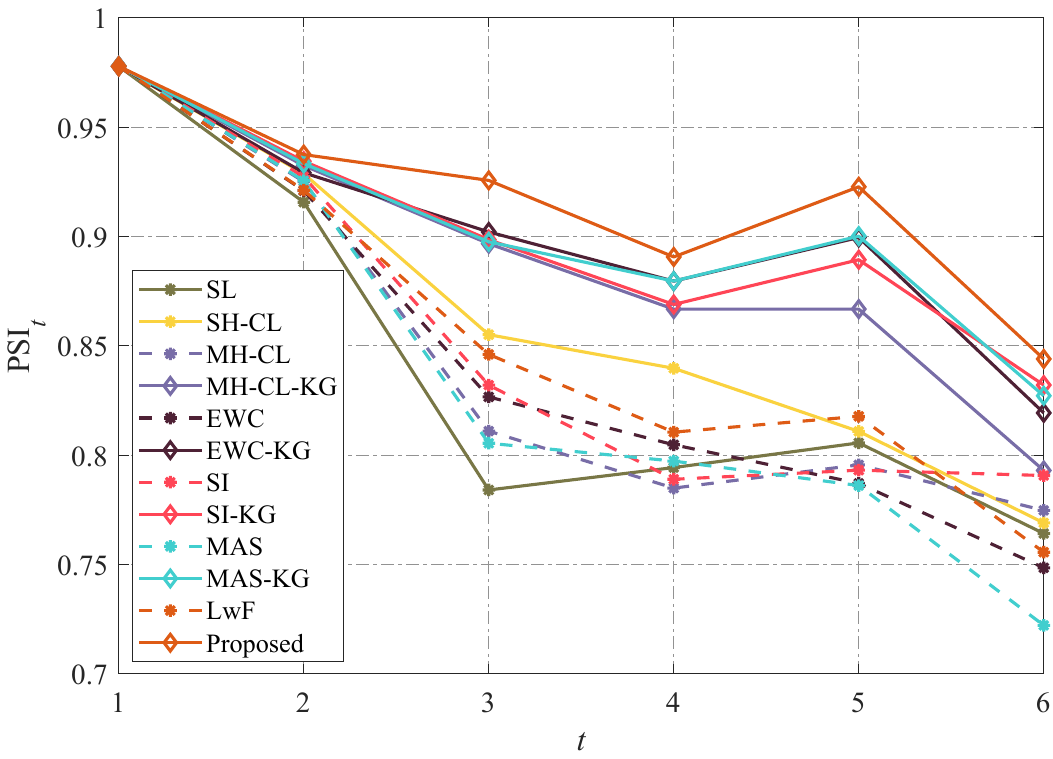}
  \caption{$\mathrm{PSI}_t$ as a function of the task index $t$.}\label{fig:psi}
\end{figure}

\begin{figure*}[ht]
    \centering
    \captionsetup{justification=centering}
    \subfloat[$\hat{q}(x) = -0.1028$ ]{\includegraphics[width=0.49\textwidth]{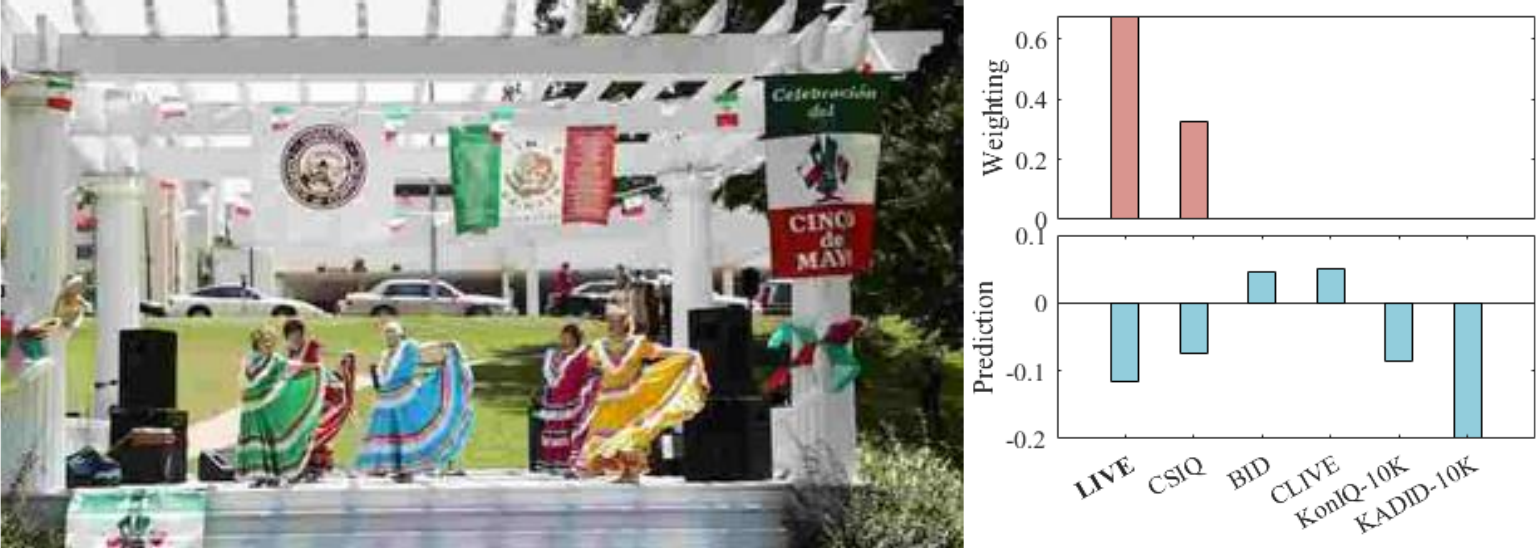}}\hskip.1em
    \subfloat[$\hat{q}(x) = -0.1940$]{\includegraphics[width=0.49\textwidth]{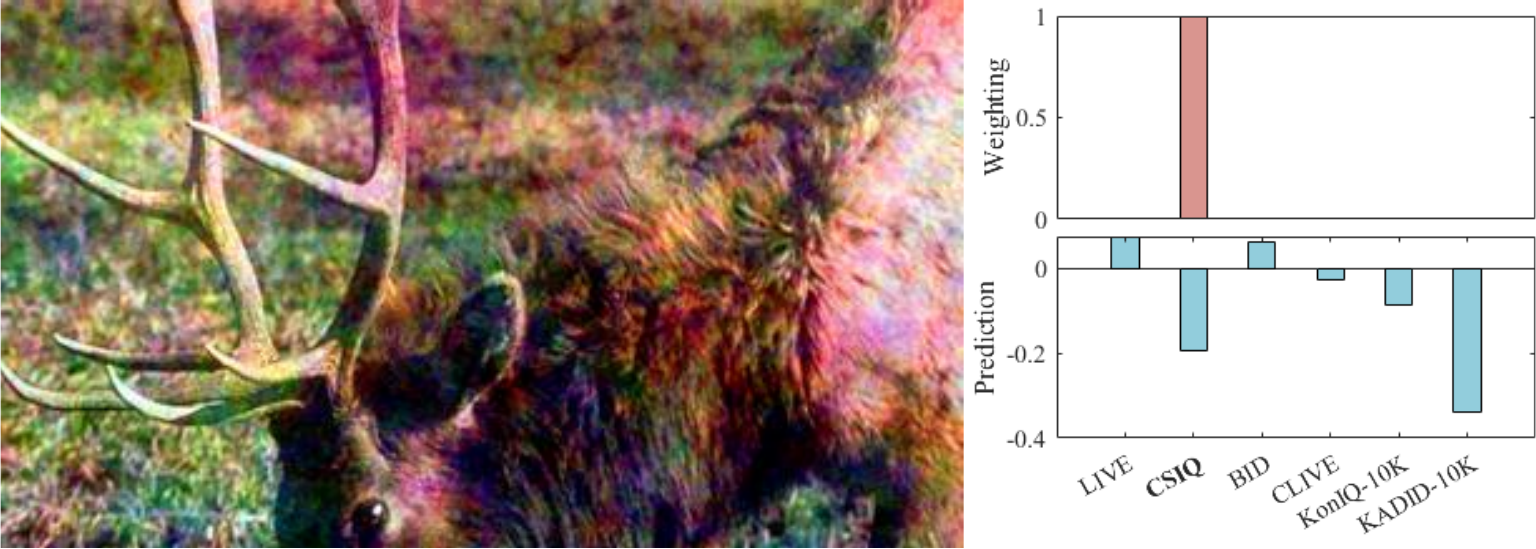}}\hskip.1em
    \subfloat[$\hat{q}(x) = 0.2624$] {\includegraphics[width=0.49\textwidth]{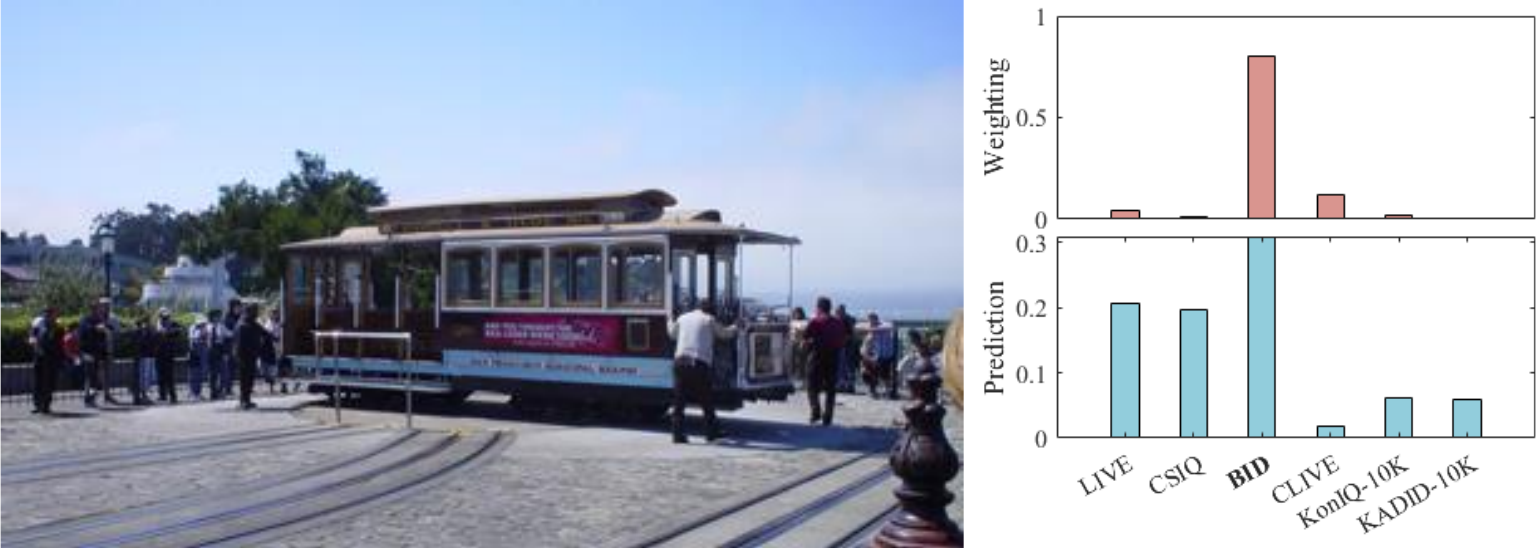}}\hskip.1em
    \subfloat[$\hat{q}(x) = 0.3736$]{\includegraphics[width=0.49\textwidth]{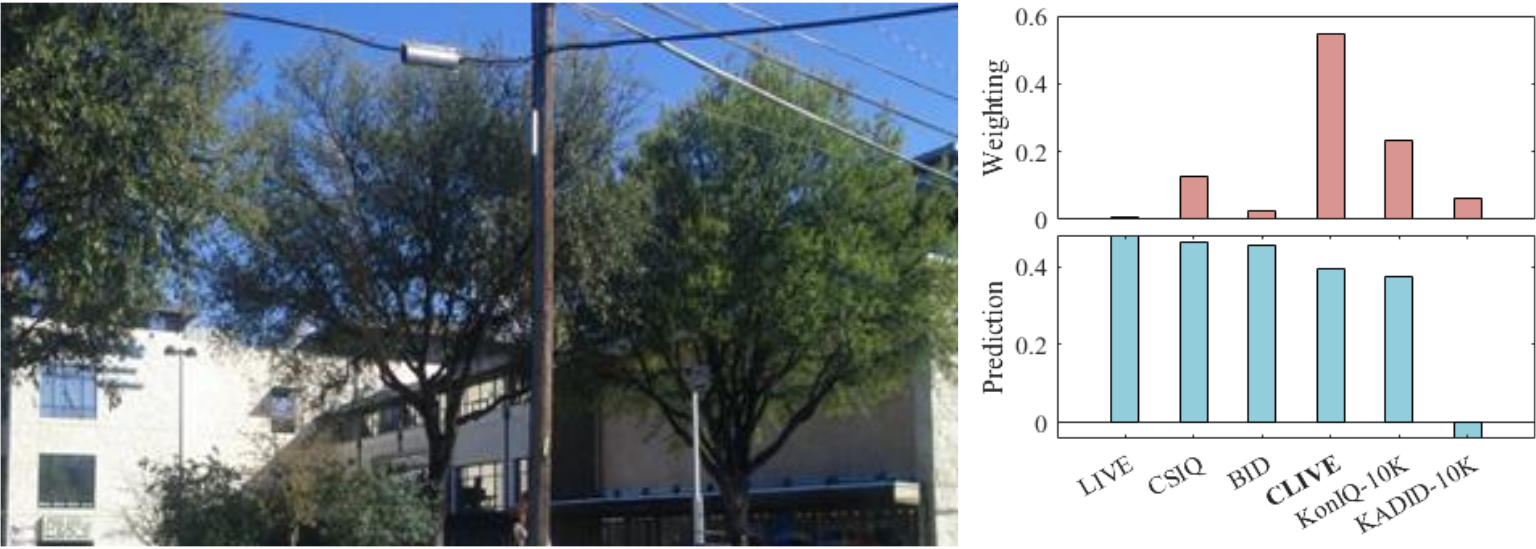}}
    \vspace{0em}
    \subfloat[$\hat{q}(x) = -0.2540$]{\includegraphics[width=0.49\textwidth]{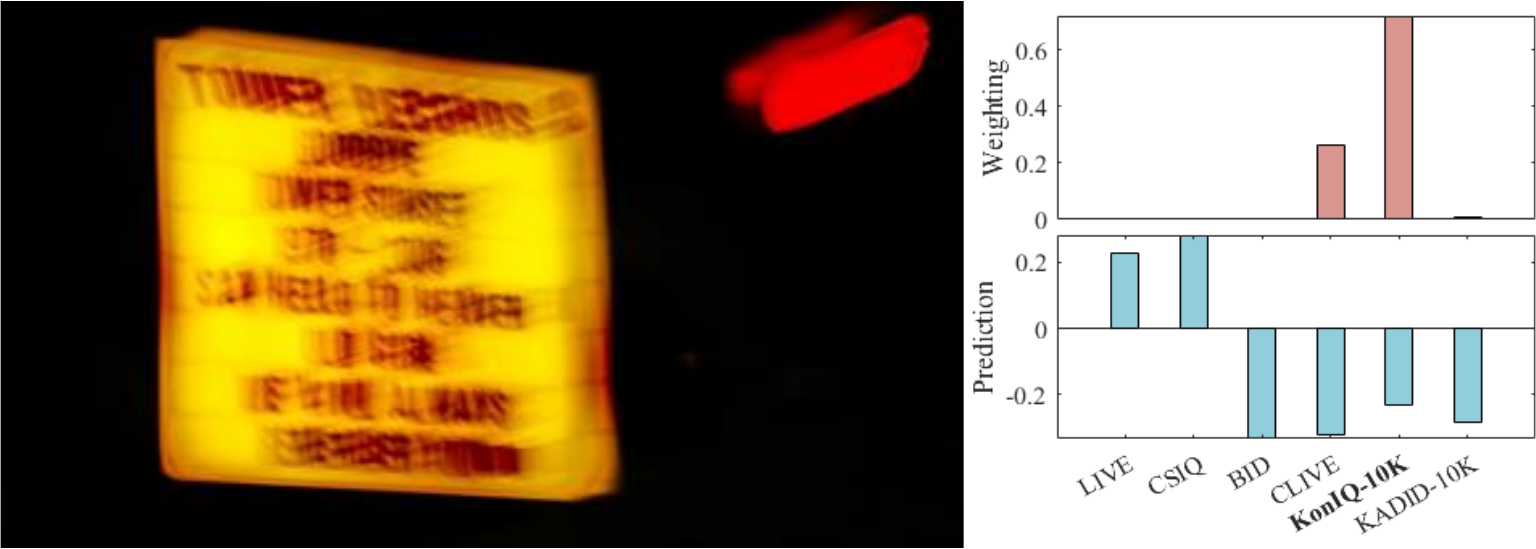}}\hskip.1em
    \subfloat[$\hat{q}(x) = -0.3009$]{\includegraphics[width=0.49\textwidth]{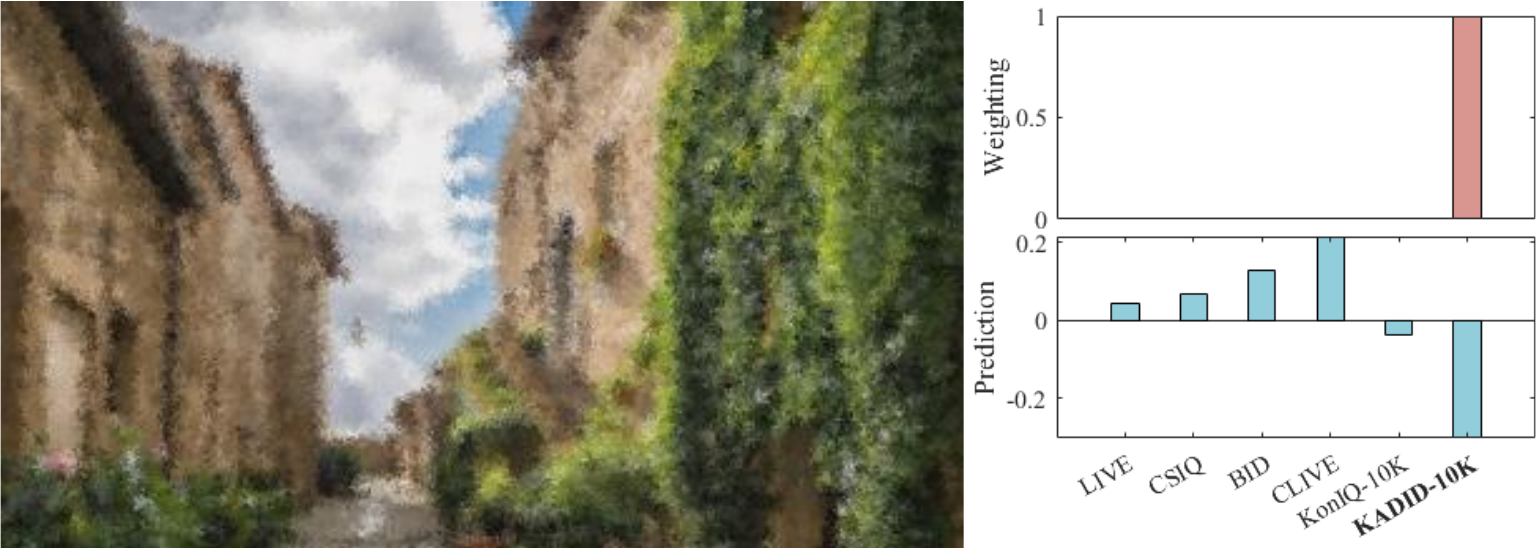}}\hskip.1em
  \caption{Perceptual scaling of images sampled from the six IQA datasets. The bar charts of weights and quality predictions of all heads are also presented alongside each image. The final quality prediction $\hat{q}(x)$ is shown in the subcaption. Zoom in for better distortion visibility.}
\label{fig:qualitative}
\end{figure*}

\subsection{Main Results}~\label{subsec:results}
In this subsection, we present the quantitative and qualitative results in the case that no previous data is directly accessible when learning new tasks.
\subsubsection{Quantitative Results}
We use the proposed $\mathrm{mSRCC}$, $\mathrm{mPI}$, $\mathrm{mSI}$, and $\mathrm{mPSI}$ in Eq.~\eqref{eq:msrcc}, Eq.~\eqref{eq:plasticity}, Eq.~\eqref{eq:stability1}, and Eq. \eqref{eq:psi} to benchmark the performance. From Table~\ref{tab:mpsr}, we have several interesting observations.
First, the unsatisfactory performance of SL calls for continual learning methods to mitigate catastrophic forgetting in BIQA. Second, SH-CL and MH-CL achieve similar $\mathrm{mPI}$ results with SL, but improves the $\mathrm{mSI}$ and $\mathrm{mSRCC}$ results upon SL by clear margins, indicating a vanilla knowledge accumulation process. Third, we see that the proposed KG mechanism leads to consistent performance gains over the baseline regularization methods. Among them, LwF-KG outperforms other regularizers in terms of $\mathrm{mSRCC}$ and $\mathrm{mPSI}$, indicating a better plasticity-stability trade-off. Fourth, when the test-time oracle is available, the performance of all methods can be further improved. We also plot $\mathrm{PSI}_{t}$ as a function of the task index $t$ in Fig.~\ref{fig:psi}, from which we find that our method is more stable, and performs much better as the length of the task sequence increases. It is noteworthy that the fluctuation of $\mathrm{PSI}_{t}$ is mainly due to the difficulty of the newly incorporated dataset and the level of subpopulation shift with respect to previous ones.

We take a closer look at the performance variations along the task sequence and summarize the SRCC results continually in Table~\ref{tab:srcc_results}. Note that all methods begin training on LIVE~\cite{sheikh2006statistical}, and their SRCC results are the same before continually learning on any new task. There are several useful findings. First, we observe that subpopulation shift between different tasks significantly oscillates the results of SL. This is not surprising because it is often challenging for BIQA models trained on datasets of synthetic distortions to perform well on datasets of realistic distortions (and vice versa)~\cite{zeng2018blind,zhang2020blind}. Second, compared with SL, SH-CL generally improves on old tasks with similar performance on new tasks, achieving a better plasticity-stability trade-off. Third, MH-CL and other regularization methods add a prediction head for each new task, which does not handle old tasks well, necessitating an effective mechanism to make full use of all learned heads. Fourth, regularization methods employing the KG module lead to better performance, especially on previous tasks. 
Compared with other regularization methods, LwF-KG delivers more stable performance. Last, all methods are upper bounded by JL as expected.

\begin{table}[!t]
  \centering
  \caption{Performance comparison of different BIQA models. All methods are trained in chronological order}\label{tab:biqa_compare}
  \begin{tabular}{lcccc}
      \toprule
      {Methods} & $\mathrm{mSRCC}$ & $\mathrm{mPI}$ & $\mathrm{mSI}$ & $\mathrm{mPSI}$\\
     \midrule
    NIQE~\cite{mittal2013making} & 0.4515 & 0.4515 & 1.0000 & 0.7258\\
    dipIQ~\cite{ma2017dipiq} & 0.3633 & 0.3633 & 1.0000 & 0.6817\\
    Ma19~\cite{ma2019blind} & 0.5664 & 0.5664 & 1.0000 & 0.7832\\
    \midrule
    BRISQUE~\cite{mittal2012no} & 0.5235 & 0.6929 & 0.6149 & 0.6539\\
    CORNIA~\cite{ye2012unsupervised} & 0.4948 & 0.6455 & 0.7069 & 0.6762\\
    DBCNN~\cite{zhang2020blind} & 0.6957 & 0.8814 & 0.7883 & 0.8349\\
    MetaIQA~\cite{zhu2020metaiqa} & 0.6841 & 0.8535 & 0.7886 & 0.8211\\
    KonCept512~\cite{hosu2020koniq} & 0.6951 & 0.8680 & 0.7853 & 0.8267\\
    \midrule
    Proposed & 0.8150 & 0.8563 & 0.9796 & 0.9180\\
     \bottomrule
  \end{tabular}
\end{table}

We find that in the continual learning setting the generalization of BIQA models is heavily influenced by the level of subpopulation shift between the seen and unseen datasets. For example, good generalization in terms of SRCC on the unseen KonIQ-10K (with realistic distortions) is achieved by the proposed method when it has been learned on BID and CLIVE with realistic distortions compared to the one trained on LIVE and CSIQ with synthetic distortions only. We also observe that the proposed method generalizes marginally to KADID-10K when it is continually trained from the 1\textsuperscript{st} to the 5\textsuperscript{th} dataset. This is because KADID-10K~\cite{lin2019kadid} contains some distinct distortion types that are not shared by previously trained datasets. In the future, we will improve the generalizability aspect of continually learned BIQA models using advanced machine learning techniques such as domain generalization~\cite{wang2021generalizing}, which is an orthogonal and complementary direction to the current work.

We also evaluate eight representative BIQA methods under the same task sequence in chronological order, including three opinion-unaware methods~\cite{mittal2013making,ma2017dipiq,ma2019blind} that do not rely on human-rated IQA datasets for training and five opinion-aware methods~\cite{mittal2012no,ye2012unsupervised,zhang2020blind,zhu2020metaiqa,hosu2020koniq} that learn from human perceptual scores. As summarized in Table~\ref{tab:biqa_compare}, we find that NIQE~\cite{mittal2013making}, dipIQ~\cite{ma2017dipiq}, and Ma19~\cite{ma2019blind} that do not learn continually deliver the maximal stability, but fail to adapt to changing distortion scenarios, resulting in poor plasticity and overall prediction accuracy. Moreover, none of the opinion-aware BIQA models is immune to catastrophic forgetting caused by the significant subpopulation shift among tasks. DNN-based models that enjoy joint optimization of feature representation and quality prediction generally exhibit better plasticity than those based on hand-engineered NSS.
 

\subsubsection{Qualitative Results}
We conduct a qualitative analysis of our LwF-KG model by sampling test images from the task sequence. Also shown in Fig.~\ref{fig:qualitative} are the bar charts of weights and quality predictions corresponding to each image. Although the proposed method is not jointly trained on all IQA datasets~\cite{zhang2020learning,zhang2021uncertainty}, it successfully learns a common perceptual scale for all tasks, within which images are well perceptually aligned. Moreover, visual inspections of the bar charts reveal that the prediction head may only give accurate quality estimates for the dataset it is exposed to. Fortunately, the proposed KG mechanism can compensate for such prediction inaccuracy, assigning larger weights to the heads trained on images with similar distortions. For example, when evaluating the images sampled from LIVE~\cite{sheikh2006statistical} (see Fig.~\ref{fig:qualitative} (a)), the head trained on CSIQ~\cite{larson2010most} of similar synthetic distortions is also assigned relatively high weighting. As another example, KADID-10K~\cite{lin2019kadid} contains some distinct distortion types (\eg, spatial jitter in Fig.~\ref{fig:qualitative} (f)); consequently, the assigned weight for the head of KADID-10K tends to dominate. In the open visual world, if a new task is difficult enough to fail the KG mechanism, we may leverage new data to improve the feature representation of the KG mechanism in the future.

\subsection{Ablation Study}\label{subsec:ablation}
In this subsection, we conduct a series of ablation experiments to evaluate the robustness of our method to different task orders and alternative design choices.

\subsubsection{Order-Robustness}\label{subsec:order}
The main experiments are conducted on the task sequence in chronological order. In real-world situations, new distortions may emerge in arbitrary order, and similar distortions may also reappear in the future. Thus a BIQA model is expected to be independent of the task order~\cite{yoon2020scalable}. To evaluate the order-robustness of the proposed method, we experiment with  four extra task orders: (\rom{1}) reverse chronological order - KADID-10K $\rightarrow$ KonIQ-10K $\rightarrow$ LIVE Challenge $\rightarrow$ BID $\rightarrow$ CSIQ $\rightarrow$ LIVE; (\rom{2}) synthetic and realistic distortions in alternation - LIVE $\rightarrow$ BID $\rightarrow$ CSIQ $\rightarrow$ LIVE Challenge $\rightarrow$ KADID-10K $\rightarrow$ KonIQ-10K; (\rom{3}) synthetic distortions followed by realistic distortions - LIVE $\rightarrow$ CSIQ $\rightarrow$ KADID-10K $\rightarrow$ BID $\rightarrow$ LIVE Challenge $\rightarrow$ KonIQ-10K; and (\rom{4}) realistic distortions followed by synthetic distortions - BID $\rightarrow$ LIVE Challenge $\rightarrow$ KonIQ-10K $\rightarrow$ LIVE $\rightarrow$ CSIQ $\rightarrow$ KADID-10K. We list the $\mathrm{mSRCC}$, $\mathrm{mPI}$, $\mathrm{mSI}$, and $\mathrm{mPSI}$ results in Table \ref{tab:results_order}, where we find our method is quite robust to handle task sequences of different orders, providing justifications for its use in real-world applications. Nevertheless, the reverse chronological order (Order \rom{1}) achieves lower $\mathrm{mPI}$, $\mathrm{mSI}$, and $\mathrm{mPSI}$ results compared to the other task orders. We believe this is because the harder task (KADID-10K) appears at the beginning of the sequence, making it more difficult for our method to trade off plasticity and stability. Our method offers an SRCC of $0.8474$ on KADID-10K~\cite{lin2019kadid} when it is first trained on, and fails to stabilize the performance with a final SRCC of $0.7703$. The weakest performance in terms of $\mathrm{mSRCC}$ is spotted in Order \rom{2}, which may be due to the frequent switching between tasks of significant subpopulation shifts (\ie, synthetic and realistic distortions).

\begin{table}[t]
  \centering
  \caption{Performance comparison for different task orders. \rom{1}: Reverse chronological order. \rom{2}: Synthetic and realistic distortions in alternation. \rom{3}: Synthetic distortions followed by  realistic distortions. \rom{4}:  Realistic distortions followed by synthetic distortions. \rom{5}: Default chronological order in bold}\label{tab:results_order}
  \begin{tabular}{lcccc}
      \toprule
      {Order} & $\mathrm{mSRCC}$ & $\mathrm{mPI}$ & $\mathrm{mSI}$ & $\mathrm{mPSI}$\\
     \midrule
\rom{1} & 0.8164 & 0.8446 & 0.9521 & 0.8984\\
    \rom{2} & 0.8049 & 0.8495 & 0.9664 & 0.9080\\
    \rom{3} & 0.8289 & 0.8617 & 0.9640 & 0.9129\\
    \rom{4} & 0.8078 & 0.8526 & 0.9633 & 0.9080\\
    \textbf{\rom{5}} & 0.8150 & 0.8563 & 0.9796 & 0.9180\\
     \bottomrule
  \end{tabular}
\end{table}

\begin{table}[t]
  \centering
  \caption{Performance comparison in terms of $\mathrm{mSRCC}$ and $\mathrm{mPSI}$ of different gating strategies}\label{tab:weighting_variant}
  \begin{tabular}{lcc}
      \toprule
      {Gating Strategy} & $\mathrm{mSRCC}$ & $\mathrm{mPSI}$ \\
     \midrule
     Nearest-Mean-of-Exemplars & 0.8081 & 0.9046 \\
     Expert Gate & 0.8090 & 0.9061 \\
    \midrule
    Without training-time KG (Proposed)& \textbf{0.8150} & \textbf{0.9180}\\
    \midrule
    With training-time KG& \textbf{0.8159} & \textbf{0.9175}\\
     \bottomrule
  \end{tabular}
\end{table}

\subsubsection{Gating Mechanism}\label{subsubsec:ablation_aw}


We compare the proposed KG mechanism with two popular expert gates in classification.  The first is the nearest-mean-of-exemplars used in iCaRL \cite{rebuffi2017icarl}, which corresponds to setting $K=1$ of the proposed KG. The only difference is that hard assignment is implemented in~\cite{rebuffi2017icarl} for classification, while soft assignment is used in our method for regression. We also implement the expert gate module in~\cite{aljundi2017expert}, which relies on autoencoder-based reconstruction~\cite{Bourlard2004AutoassociationBM}. Our autoencoder consists of two fully-connected layers, one for encoding and the other for decoding, respectively, with ReLU in between. During inference, the weights are computed according to the reconstruction errors through a softmin function. The $\mathrm{mSRCC}$ and $\mathrm{mPSI}$ results are listed in Table~\ref{tab:weighting_variant}, where we see that the proposed KG clearly outperforms the nearest-mean-of-exemplars and the expert gate. We believe this arises because the intra-variance of each dataset in BIQA (containing different distortion types at varying degradation levels) is relatively large, requiring multiple distortion-aware centroids for task summarization.
 
The forward computation of the network is different between training and inference time when the KG mechanism is used. We conduct another set of experiments to see whether incorporation of the KG mechanism during training brings additional performance gains. As shown in Table~\ref{tab:weighting_variant}, we observe similar performance in terms of $\mathrm{mSRCC}$ and $\mathrm{mPSI}$ with and without the training-time KG mechanism, while keeping other training procedures the same.
 
We also take the index of the maximum weight in KG as the prediction of the dataset the test image belongs to. We compare the confusion matrices of the pre-trained ResNet-18 and the VGG-like CNN in Table~\ref{tab:confusion_matrix}. The VGG-like distortion-aware representation achieves much higher gating accuracies, which is directly reflected in quality prediction improvements.

\begin{table}[t]
  \centering
  \caption{Confusion matrices of the KG mechanism with different feature representations}\label{tab:confusion_matrix}
  \begin{tabular}{l|cccccc}
      \toprule
    Feature   & \multicolumn{6}{c}{Pre-trained ResNet-18}\\\cline{2-7}\hline
    Dataset & LIVE & CSIQ & BID & CLIVE & KonIQ& KADID\\\hline
     LIVE & 0.5313 & 0.0813 & 0.0000 & 0.2438 & 0.0750 & 0.0688\\
     CSIQ & 0.0523 & 0.2791 & 0.0349 & 0.1453 & 0.3953 & 0.0930\\
     BID & 0.0000 & 0.0085 & 0.6496 & 0.1795 & 0.1624 & 0.0000\\
     CLIVE & 0.0043 & 0.0089 & 0.1202 & 0.7210 & 0.1459 & 0.0000\\
     KonIQ & 0.0020 & 0.0010 & 0.0581 & 0.0407 & 0.8918 & 0.0065 \\
     KADID & 0.0520 & 0.0355 & 0.0280 & 0.2750 & 0.2175 &  0.3920\\
      \midrule
   Feature   & \multicolumn{6}{c}{VGG-like CNN}\\\cline{2-7}\hline
    Dataset   & LIVE & CSIQ & BID & CLIVE & KonIQ& KADID\\
     \hline
     LIVE & \textbf{0.7563} & 0.1438 & 0.0063 & 0.0125 & 0.0063 & 0.0750\\
     CSIQ & 0.1919 & \textbf{0.5988} & 0.0000 & 0.0814 & 0.0116 & 0.1163\\
     BID & 0.0085 & 0.0000 & \textbf{0.9231} & 0.0256 & 0.0427 & 0.0000\\
     CLIVE & 0.0215 & 0.0258 & 0.0410 & \textbf{0.8884} & 0.0300 & 0.0129\\
     KonIQ & 0.0015 & 0.0074 & 0.0134 & 0.0362 & \textbf{0.9236} & 0.0179\\
     KADID & 0.0545 & 0.0700 & 0.0165 & 0.0160 & 0.0365 & \textbf{0.8065}\\
     \bottomrule
  \end{tabular}
\end{table}

\begin{table}[t]
  \centering
  \caption{Performance comparison of different feature representations}\label{tab:backbone_compare}
  \begin{tabular}{lcccc}
      \toprule
      {Feature Representation} & $\mathrm{mSRCC}$ & $\mathrm{mPI}$ & $\mathrm{mSI}$ & $\mathrm{mPSI}$\\
     \midrule
    Single Stream-GAP & 0.7804 & 0.7776 & 0.9514 & 0.8645\\
    Single Stream-BP & 0.8092 & 0.8267 & \textbf{0.9825} & 0.9046\\
    Two Stream-BP & \textbf{0.8150} & \textbf{0.8563} & 0.9796 & \textbf{0.9180}\\
     \bottomrule
  \end{tabular}
\end{table}

\subsubsection{Feature Representation}\label{subsubsec:representation}
We compare the adopted two-stream network with two variants:  a single-stream ResNet-18 network with 1) global average pooling and 2)  bilinear pooling as feature extractors, respectively, denoted by \textbf{Single Stream-GAP} and \textbf{Single Stream-BP}. Note that both variants also use the VGG-like CNN to perform KG during inference, yet its features are ablated from the quality representation. 
From Table~\ref{tab:backbone_compare}, we observe that Single Stream-BP consistently outperforms Single Stream-GAP for all metrics. The two-stream backbone further boosts the model plasticity, leading to higher $\mathrm{mSRCC}$ and $\mathrm{mPSI}$ results.


\subsection{Further Results based on Experience Replay}\label{subsec:replay}
In this subsection, we evaluate different continual learning methods based on experience replay for BIQA, where previous data is partially accessible.

\subsubsection{Memory Management}\label{subsubsec:memory_manage}
The very first step in experience replay is to manage the memory budget. Inspired by~\cite{prabhu2020gdumb}, we assume all tasks and all data in each task to be equally important. We greedily update a memory buffer $\mathcal{M}$ to accommodate new data while keeping a balanced task distribution, as presented in Algorithm~\ref{algo}.

\begin{algorithm}[t]
\caption{Memory Management}
\label{algo}
{\bf Input:}
$\{\mathcal{D}_t\}_{t=1}^{T} = \{\{x^{(i)}_t, \mu^{(i)}_t\}_{i=1}^{\vert\mathcal{D}_t\vert}\}_{t=1}^{T}$\\
{\bf Require:}
Memory buffer: $\mathcal{M}$, memory budget: $B$, task length: $T$, and $\mathrm{RandSample}(\mathcal{D}, b)$: randomly sample $b$ images from $\mathcal{D}$
\begin{algorithmic}[1]
\State $\mathcal{M} \leftarrow \emptyset$
\State $b = \min(B, \vert\mathcal{D}_{1}\vert)$
\State  $\mathcal{M}_1 = \mathrm{RandSample}(\mathcal{D}_{1}, b)$
\State $\mathcal{M} = \mathcal{M}\bigcup \mathcal{M}_1$
\For{$t = 2, \ldots, T$} 
\State $\mathcal{M} \leftarrow \emptyset$
        \State $m = \lfloor\frac{{ B}}{t}\rfloor$ \quad $\triangleright$ Divide the memory budget
        \For{$k = 1, \ldots, t-1$}
            \State $b = \min(m, \vert\mathcal{M}_{k}\vert)$
            \State  $\mathcal{M}_k \leftarrow \mathcal{M}_k \setminus \mathrm{RandSample}(\mathcal{M}_{k}, \vert\mathcal{M}_{k}\vert - b)$
        \EndFor
        \State $b = \min(m, \vert\mathcal{D}_{t}\vert)$
\State  $\mathcal{M}_t \leftarrow \mathrm{RandSample}(\mathcal{D}_{t}, b)$
\State $\mathcal{M}= \bigcup_{k=1}^t\mathcal{M}_k$
\EndFor
{\bf end for}
\end{algorithmic}
\end{algorithm}

\begin{table*}[t]
  \centering
  \caption{Performance comparison in terms of $\mathrm{mSRCC}$, $\mathrm{mPI}$, $\mathrm{mSI}$, and $\mathrm{mPSI}$ of different experience replay methods with varying memory budgets. Results of the proposed LwF-KG are listed as reference}\label{tab:er_results}
  \begin{tabular}{l|ccccc|ccccc}
      \toprule
        Memory Budget  & 100 & 500 & 1,000 & 2,000 & 5,000  & 100  & 500 & 1,000 & 2,000 & 5,000\\
    \hline
      Measure   & \multicolumn{5}{c|}{$\mathrm{mSRCC}$} & \multicolumn{5}{c}{$\mathrm{mPI}$}\\
    \hline
     LwF-KG   & \multicolumn{5}{c|}{0.8150} & \multicolumn{5}{c}{0.8563}\\
     \hline
        SH-CL-ER  & \textbf{0.8249} & \textbf{0.8601} & 0.8687 & 0.8734 & 0.8736 & \textbf{0.8795} & \textbf{0.8844} & \textbf{0.8801} & \textbf{0.8802} & \textbf{0.8793} \\
        MH-CL-ER  & 0.7970 & 0.8256 & 0.8571 & 0.8572 & 0.8658 & 0.8713 & 0.8717 & 0.8730 & 0.8744 & 0.8759\\
        iCaRL-v1 & 0.8021 & 0.8369 & 0.8534 & 0.8603 & 0.8704 & 0.8662 & 0.8543 & 0.8640 & 0.8529 & 0.8721\\
        iCaRL-v2 & 0.7998 & 0.8452 & \textbf{0.8712} & \textbf{0.8746} & \textbf{0.8844} & 0.8729 & 0.8560 & 0.8672 & 0.8660 & 0.8775 \\
        GDumb  & 0.5669 & 0.7823 & 0.7978 & 0.8188 & 0.8445 & 0.6971 & 0.7883 & 0.8226 & 0.8507 & 0.8676 \\
    \midrule
        Memory Budget  & 100 & 500 & 1,000 & 2,000 & 5,000 & 100 & 500 & 1,000 & 2,000 & 5,000\\
    \hline
      Measure   & \multicolumn{5}{c|}{$\mathrm{mSI}$} & \multicolumn{5}{c}{$\mathrm{mPSI}$}\\
     \hline
     LwF-KG   & \multicolumn{5}{c|}{0.9796} & \multicolumn{5}{c}{0.9180}\\
     \hline
        SH-CL-ER  & 0.9282 & 0.9487 & 0.9445 & 0.9454 & 0.9459 &  0.9036 & 0.9166 & 0.9123 & 0.9128 & 0.9125 \\
        MH-CL-ER  & 0.9420 & 0.9648 & 0.9686 & 0.9742 & \textbf{0.9753} &  0.9067 & \textbf{0.9183} & 0.9208 & \textbf{0.9243} & \textbf{0.9256} \\
        iCaRL-v1  & \textbf{0.9548} & \textbf{0.9743} & \textbf{0.9810} & \textbf{0.9789} & 0.9735  & \textbf{0.9105} & 0.9143 & 0.9225 & 0.9159 & 0.9228 \\
        iCaRL-v2  & 0.9451 & 0.9697 & 0.9747 & 0.9783 & 0.9638  & 0.9090 & 0.9129 & \textbf{0.9230} & 0.9222 & 0.9207 \\
        GDumb & 0.9226 & 0.9586 & 0.9538 & 0.9474 & 0.9517 & 0.8099 & 0.8735 & 0.8882 & 0.8991 & 0.9097 \\  
     \bottomrule
  \end{tabular}
\end{table*}

\subsubsection{Competing Methods with Experience Replay}\label{subsubsec:rehearsal_cl}
Under the pairwise learning-to-rank framework, we need to transform $\mathcal{M}$ to $\mathcal{P}_{\mathcal{M}}$ following the procedure described in Subsubsection~\ref{subsubsec:d1}. We consider five experience replay methods.\\
     \textbf{SH-CL with Experience Replay}  (\textbf{SH-CL-ER}) is built on SH-CL that trains a single-head model using mini-batches of images sampled from both $\mathcal{P}_{t}$ and $\mathcal{P}_\mathcal{M}$.\\
      \textbf{MH-CL with Experience Replay} (\textbf{MH-CL-ER}) updates each old head using the corresponding images from $\mathcal{P}_\mathcal{M}$ and the current head using $\mathcal{P}_{t}$, respectively. KG is used for quality prediction.\\
      \textbf{iCaRL-v1} is a direct adaptation of iCaRL~\cite{rebuffi2017icarl} in image classification. Driven by knowledge distillation, iCaRL-v1 uses the predictions of the $k$-th head to compute a probability $\bar{p}_{tk}(x,y)$ as the pseudo label for $(x,y)\in \mathcal{M}_k$, and trains the model using the fidelity loss.
      The main difference between iCaRL-v1 and LwF is that they distill knowledge using image samples from the memory buffer $\mathcal{M}$ and the current dataset $\mathcal{D}_{t}$, respectively.\\
     \textbf{iCaRL-v2} is an advanced adaptation of iCaRL to BIQA. During learning from the $t$-th task, it performs the same as iCaRL-v1 for all previous heads, and optimizes the current head using the combination of $\mathcal{P}_t$ and $\mathcal{P}_\mathcal{M}$ (as in SH-CL-ER). By doing so, the current head is also exposed to a portion of data from previous tasks, aiming for more accurate predictions. This motivates us to compute the final quality score by first aggregating predictions from old tasks with KG and then averaging it with the prediction of the current head.\\
     \textbf{GDumb} is a simple baseline~\cite{prabhu2020gdumb} that trains a model with the memory buffer $\mathcal{M}$ only. Although GDumb is not specifically designed for continual learning, we include it in our experiments because GDumb performs competitively against many continual learning methods in image classification~\cite{mai2021online}.

\subsubsection{Results and Analysis}\label{subsubsec:rehearsal_results}
We compare the performance of different experience replay schemes in Table \ref{tab:er_results}, where we organize the tasks in chronological order. The key observation is that maintaining a memory buffer with a reasonable budget (\eg, $B$=1,000) leads to noticeable performance improvements over the reference model LwF-KG, especially under $\mathrm{mSRCC}$. Nevertheless, small memory budgets (\eg, $B$=100) may hurt the prediction accuracy and the plasticity-stability trade-off due to the danger of over-fitting  $\mathcal{M}$. In other words, if the memory budget is extremely limited, replay-free continual learning methods (\eg, the proposed LwF-KG) may be preferable. Incorporating the learning strategy of SH-CL-ER, iCaRL-v2 shows stronger plasticity than iCaRL-v1, while the latter exhibits stronger stability. For $B=$1,000, iCaRL-v2 is the best performer in terms of both $\mathrm{mSRCC}$ and $\mathrm{mPSI}$. Despite the remarkable performance in image classification, GDumb appears much more data-hungry in BIQA.

\section{Conclusion}
We have formulated continual learning for BIQA with five desiderata and three performance measures. We also contributed continual learning methods to train BIQA models robust to subpopulation shift in this new setting.

This work establishes a new research direction in BIQA with many important topics left unexplored. First, it remains wide open whether we need to add or remove several desiderata to make continual learning for BIQA more practical. For example, it may be useful to add the online learning desideratum, where learning happens instantaneously with no distinct boundaries between tasks (or datasets). Second, better continual learning methods for BIQA are desirable to bridge the performance gap between the current method and the upper bound by joint learning. Third, with access to partial data of previous tasks, better experience replay strategies would be valuable directions to pursue. Fourth, the current work only considers two distortion scenarios, \ie, synthetic and realistic distortions, to construct the task sequence. In the future, it would be interesting to incorporate multiple distortion scenarios, representing severer subpopulation shift during training and testing. Last, the current work only explores small-length task sequences with a limited number of task orders. It is necessary to test the current method on task sequences with arbitrary length and in arbitrary order. It is also important to develop more order-robust and length-robust continual learning methods for BIQA. 

\ifCLASSOPTIONcompsoc
  \section*{Acknowledgments}
\else
  \section*{Acknowledgment}
\fi
This work was supported in part by the National Natural Science Foundation of China under Grants 61901262, 62071407, and U19B2035, Shanghai Municipal Science and Technology Major Project (2021SHZDZX0102), and  the Hong Kong RGC Early Career Scheme (9048212). The authors would like to sincerely thank the three anonymous reviewers in the first round for their highest-quality comments, which significantly improve the manuscript. The authors would also like to thank Xuelin Liu for helping illustrate the idea of subpopulation shift in Fig.~\ref{fig:illustration}. 

\bibliographystyle{IEEEtran}
\bibliography{Weixia}

\begin{thebibliography}{10}
\providecommand{\url}[1]{#1}
\csname url@samestyle\endcsname
\providecommand{\newblock}{\relax}
\providecommand{\bibinfo}[2]{#2}
\providecommand{\BIBentrySTDinterwordspacing}{\spaceskip=0pt\relax}
\providecommand{\BIBentryALTinterwordstretchfactor}{4}
\providecommand{\BIBentryALTinterwordspacing}{\spaceskip=\fontdimen2\font plus
\BIBentryALTinterwordstretchfactor\fontdimen3\font minus
  \fontdimen4\font\relax}
\providecommand{\BIBforeignlanguage}[2]{{%
\expandafter\ifx\csname l@#1\endcsname\relax
\typeout{** WARNING: IEEEtran.bst: No hyphenation pattern has been}%
\typeout{** loaded for the language `#1'. Using the pattern for}%
\typeout{** the default language instead.}%
\else
\language=\csname l@#1\endcsname
\fi
#2}}
\providecommand{\BIBdecl}{\relax}
\BIBdecl

\bibitem{wang2006modern}
Z.~Wang and A.~C. Bovik, \emph{Modern Image Quality Assessment}.\hskip 1em plus
  0.5em minus 0.4em\relax Morgan \& Claypool, 2006.

\bibitem{sheikh2006statistical}
H.~R. Sheikh, M.~F. Sabir, and A.~C. Bovik, ``A statistical evaluation of
  recent full reference image quality assessment algorithms,'' \emph{IEEE
  Transactions on Image Processing}, vol.~15, no.~11, pp. 3440--3451, Nov.
  2006.

\bibitem{wang2002no}
Z.~Wang, H.~R. Sheikh, and A.~C. Bovik, ``No-reference perceptual quality
  assessment of {JPEG} compressed images,'' in \emph{IEEE International
  Conference on Image Processing}, vol.~1, 2002, pp. 477--480.

\bibitem{mittal2012no}
A.~Mittal, A.~K. Moorthy, and A.~C. Bovik, ``No-reference image quality
  assessment in the spatial domain,'' \emph{IEEE Transactions on Image
  Processing}, vol.~21, no.~12, pp. 4695--4708, Dec. 2012.

\bibitem{ye2012unsupervised}
P.~Ye, J.~Kumar, L.~Kang, and D.~Doermann, ``Unsupervised feature learning
  framework for no-reference image quality assessment,'' in \emph{IEEE
  Conference on Computer Vision and Pattern Recognition}, 2012, pp. 1098--1105.

\bibitem{larson2010most}
E.~C. Larson and D.~M. Chandler, ``Most apparent distortion: {F}ull-reference
  image quality assessment and the role of strategy,'' \emph{Journal of
  Electronic Imaging}, vol.~19, no.~1, pp. 1--21, Jan. 2010.

\bibitem{ponomarenko2013color}
P.~Nikolay, J.~Lina, I.~Oleg, L.~Vladimir, E.~Karen, A.~Jaakko, V.~Benoit,
  C.~Kacem, C.~Marco, B.~Federica, and C.-C.~J. Kuo, ``Image database
  {TID2013}: Peculiarities, results and perspectives,'' \emph{Signal
  Processing: Image Communication}, vol.~30, pp. 57--77, Jan. 2015.

\bibitem{lin2019kadid}
H.~Lin, V.~Hosu, and D.~Saupe, ``K{ADID}-10k: A large-scale artificially
  distorted {IQA} database,'' in \emph{International Conference on Quality of
  Multimedia Experience}, 2019, pp. 1--3.

\bibitem{ma2017waterloo}
K.~Ma, Z.~Duanmu, Q.~Wu, Z.~Wang, H.~Yong, H.~Li, and L.~Zhang, ``Waterloo
  {E}xploration {D}atabase: New challenges for image quality assessment
  models,'' \emph{IEEE Transactions on Image Processing}, vol.~26, no.~2, pp.
  1004--1016, Feb. 2017.

\bibitem{ghadiyaram2016massive}
D.~Ghadiyaram and A.~C. Bovik, ``Massive online crowdsourced study of
  subjective and objective picture quality,'' \emph{IEEE Transactions on Image
  Processing}, vol.~25, no.~1, pp. 372--387, Jan. 2016.

\bibitem{hosu2020koniq}
V.~Hosu, H.~Lin, T.~Sziranyi, and D.~Saupe, ``Kon{IQ}-10k: An ecologically
  valid database for deep learning of blind image quality assessment,''
  \emph{IEEE Transactions on Image Processing}, vol.~29, pp. 4041--4056, Jan.
  2020.

\bibitem{fang2020perceptual}
Y.~Fang, H.~Zhu, Y.~Zeng, K.~Ma, and Z.~Wang, ``Perceptual quality assessment
  of smartphone photography,'' in \emph{IEEE Conference on Computer Vision and
  Pattern Recognition}, 2020, pp. 3677--3686.

\bibitem{bosse2016deep}
S.~Bosse, D.~Maniry, K.~R. M{\"u}ller, T.~Wiegand, and W.~Samek, ``Deep neural
  networks for no-reference and full-reference image quality assessment,''
  \emph{IEEE Transactions on Image Processing}, vol.~27, no.~1, pp. 206--219,
  Jan. 2018.

\bibitem{Ma2018End}
K.~Ma, W.~Liu, K.~Zhang, Z.~Duanmu, Z.~Wang, and W.~Zuo, ``End-to-end blind
  image quality assessment using deep neural networks,'' \emph{IEEE
  Transactions on Image Processing}, vol.~27, no.~3, pp. 1202--1213, Mar. 2018.

\bibitem{mittal2013making}
A.~Mittal, R.~Soundararajan, and A.~C. Bovik, ``Making a `completely blind'
  image quality analyzer,'' \emph{IEEE Signal Processing Letters}, vol.~20,
  no.~3, pp. 209--212, Mar. 2013.

\bibitem{zhang2015feature}
L.~Zhang, L.~Zhang, and A.~C. Bovik, ``A feature-enriched completely blind
  image quality evaluator,'' \emph{IEEE Transactions on Image Processing},
  vol.~24, no.~8, pp. 2579--2591, Aug. 2015.

\bibitem{mccloskey1989catastrophic}
M.~McCloskey and N.~J. Cohen, ``Catastrophic interference in connectionist
  networks: The sequential learning problem,'' in \emph{Psychology of Learning
  and Motivation}, 1989, vol.~24, pp. 109--165.

\bibitem{zhang2020learning}
W.~Zhang, K.~Ma, G.~Zhai, and X.~Yang, ``Learning to blindly assess image
  quality in the laboratory and wild,'' in \emph{IEEE International Conference
  on Image Processing}, 2020, pp. 111--115.

\bibitem{zhang2021uncertainty}
------, ``Uncertainty-aware blind image quality assessment in the laboratory
  and wild,'' \emph{IEEE Transactions on Image Processing}, vol.~30, pp.
  3474--3486, Mar. 2021.

\bibitem{aljundi2019continual}
R.~Aljundi, ``Continual learning in neural networks,'' \emph{CoRR}, vol.
  abs/1910.02718, 2019.

\bibitem{ciancio2011no}
A.~Ciancio, A.~L. N.~T. {Targino da Costa}, E.~A.~B. {da Silva}, A.~Said,
  R.~Samadani, and P.~Obrador, ``No-reference blur assessment of digital
  pictures based on multifeature classifiers,'' \emph{IEEE Transactions on
  Image Processing}, vol.~20, no.~1, pp. 64--75, Jan. 2011.

\bibitem{rebuffi2017icarl}
S.-A. Rebuffi, A.~Kolesnikov, G.~Sperl, and C.~H. Lampert, ``{iCaRL}:
  Incremental classifier and representation learning,'' in \emph{IEEE
  Conference on Computer Vision and Pattern Recognition}, 2017, pp. 2001--2010.

\bibitem{ying2020from}
Z.~Ying, H.~Niu, P.~Gupta, D.~Mahajan, D.~Ghadiyaram, and A.~Bovik, ``From
  patches to pictures ({P}a{Q}-2-{P}i{Q}): Mapping the perceptual space of
  picture quality,'' in \emph{IEEE Conference on Computer Vision and Pattern
  Recognition}, 2020, pp. 3575--3585.

\bibitem{marziliano2004perceptual}
P.~Marziliano, F.~Dufaux, S.~Winkler, and T.~Ebrahimi, ``Perceptual blur and
  ringing metrics: application to {JPEG}2000,'' \emph{Signal Processing: Image
  Communication}, vol.~19, no.~2, pp. 163--172, Feb. 2004.

\bibitem{moorthy2011blind}
A.~K. Moorthy and A.~C. Bovik, ``Blind image quality assessment: From natural
  scene statistics to perceptual quality,'' \emph{IEEE Transactions on Image
  Processing}, vol.~20, no.~12, pp. 3350--3364, Dec. 2011.

\bibitem{saad2012blind}
M.~A. Saad, A.~C. Bovik, and C.~Charrier, ``Blind image quality assessment: A
  natural scene statistics approach in the {DCT} domain,'' \emph{IEEE
  Transactions on Image Processing}, vol.~21, no.~8, pp. 3339--3352, Aug. 2012.

\bibitem{simoncelli2001natural}
E.~P. Simoncelli and B.~A. Olshausen, ``Natural image statistics and neural
  representation,'' \emph{Annual Review of Neuroscience}, vol.~24, no.~1, pp.
  1193--1216, Aug. 2001.

\bibitem{xu2016blind}
J.~Xu, P.~Ye, Q.~Li, H.~Du, Y.~Liu, and D.~Doermann, ``Blind image quality
  assessment based on high order statistics aggregation,'' \emph{IEEE
  Transactions on Image Processing}, vol.~25, no.~9, pp. 4444--4457, Sep. 2016.

\bibitem{ghadiyaram2017perceptual}
D.~Ghadiyaram and A.~C. Bovik, ``Perceptual quality prediction on authentically
  distorted images using a bag of features approach,'' \emph{Journal of
  Vision}, vol.~17, no.~1, pp. 32--32, Jan. 2017.

\bibitem{kang2014convolutional}
L.~Kang, P.~Ye, Y.~Li, and D.~Doermann, ``Convolutional neural networks for
  no-reference image quality assessment,'' in \emph{IEEE Conference on Computer
  Vision and Pattern Recognition}, 2014, pp. 1733--1740.

\bibitem{zeng2018blind}
H.~Zeng, L.~Zhang, and A.~C. Bovik, ``Blind image quality assessment with a
  probabilistic quality representation,'' in \emph{IEEE International
  Conference on Image Processing}, 2018, pp. 609--613.

\bibitem{liu2017rankiqa}
X.~Liu, J.~v.~d. Weijer, and A.~D. Bagdanov, ``{RankIQA}: Learning from
  rankings for no-reference image quality assessment,'' in \emph{IEEE
  International Conference on Computer Vision}, 2017, pp. 1040--1049.

\bibitem{zhang2020blind}
W.~Zhang, K.~Ma, J.~Yan, D.~Deng, and Z.~Wang, ``Blind image quality assessment
  using a deep bilinear convolutional neural network,'' \emph{IEEE Transactions
  on Circuits and Systems for Video Technology}, vol.~30, no.~1, pp. 36--47,
  Jan. 2020.

\bibitem{ma2019blind}
K.~Ma, X.~Liu, Y.~Fang, and E.~P. Simoncelli, ``Blind image quality assessment
  by learning from multiple annotators,'' in \emph{IEEE International
  Conference on Imaging Processing}, 2019, pp. 2344--2348.

\bibitem{9121773}
J.~Wu, J.~Ma, F.~Liang, W.~Dong, G.~Shi, and W.~Lin, ``End-to-end blind image
  quality prediction with cascaded deep neural network,'' \emph{IEEE
  Transactions on Image Processing}, vol.~29, pp. 7414--7426, Jun. 2020.

\bibitem{wang2020active}
Z.~Wang and K.~Ma, ``Active fine-tuning from {gMAD} examples improves blind
  image quality assessment,'' \emph{IEEE Transactions on Pattern Analysis and
  Machine Intelligence}, to appear, 2021.

\bibitem{zhu2020metaiqa}
H.~Zhu, L.~Li, J.~Wu, W.~Dong, and G.~Shi, ``Meta{IQA}: Deep meta-learning for
  no-reference image quality assessment,'' in \emph{IEEE Conference on Computer
  Vision and Pattern Recognition}, 2020, pp. 14\,131--14\,140.

\bibitem{li2020norm}
D.~Li, T.~Jiang, and M.~Jiang, ``Norm-in-norm loss with faster convergence and
  better performance for image quality assessment,'' in \emph{ACM International
  Conference on Multimedia}, 2020, pp. 789--797.

\bibitem{9156687}
S.~Su, Q.~Yan, Y.~Zhu, C.~Zhang, X.~Ge, J.~Sun, and Y.~Zhang, ``Blindly assess
  image quality in the wild guided by a self-adaptive hyper network,'' in
  \emph{IEEE Conference on Computer Vision and Pattern Recognition}, 2020, pp.
  3664--3673.

\bibitem{french1999catastrophic}
R.~M. French, ``Catastrophic forgetting in connectionist networks,''
  \emph{Trends in Cognitive Sciences}, vol.~3, no.~4, pp. 128--135, Apr. 1999.

\bibitem{li2017learning}
Z.~Li and D.~Hoiem, ``Learning without forgetting,'' \emph{IEEE Transactions on
  Pattern Analysis and Machine Intelligence}, vol.~40, no.~12, pp. 2935--2947,
  Dec. 2017.

\bibitem{hinton2015distilling}
G.~Hinton, O.~Vinyals, and J.~Dean, ``Distilling the knowledge in a neural
  network,'' \emph{CoRR}, vol. abs/1503.02531, 2015.

\bibitem{rannen2017encoder}
A.~Rannen, R.~Aljundi, M.~B. Blaschko, and T.~Tuytelaars, ``Encoder based
  lifelong learning,'' in \emph{IEEE International Conference on Computer
  Vision}, 2017, pp. 1320--1328.

\bibitem{aljundi2017expert}
R.~Aljundi, P.~Chakravarty, and T.~Tuytelaars, ``Expert gate: Lifelong learning
  with a network of experts,'' in \emph{IEEE Conference on Computer Vision and
  Pattern Recognition}, 2017, pp. 3366--3375.

\bibitem{kirkpatrick2017overcoming}
J.~Kirkpatrick, R.~Pascanu, N.~Rabinowitz, J.~Veness, G.~Desjardins, A.~A.
  Rusu, K.~Milan, Q.~John, T.~Ramalho, A.~Grabska-Barwinska, C.~Clopath,
  D.~Kumaran, and R.~Hadsell, ``Overcoming catastrophic forgetting in neural
  networks,'' \emph{Proceedings of the National Academy of Sciences}, vol. 114,
  no.~13, pp. 3521--3526, Mar. 2017.

\bibitem{schwarz2018progress}
J.~Schwarz, W.~Czarnecki, J.~Luketina, A.~Grabska-Barwinska, Y.~W. Teh,
  R.~Pascanu, and R.~Hadsell, ``Progress \& compress: A scalable framework for
  continual learning,'' in \emph{International Conference on Machine Learning},
  2018, pp. 4528--4537.

\bibitem{lee2017overcoming}
S.-W. Lee, J.-H. Kim, J.~Jun, J.-W. Ha, and B.-T. Zhang, ``Overcoming
  catastrophic forgetting by incremental moment matching,'' in \emph{Advances
  in Neural Information Processing Systems}, 2017, pp. 4652--4662.

\bibitem{nguyen2018variational}
C.~V. Nguyen, Y.~Li, T.~D. Bui, and R.~E. Turner, ``Variational continual
  learning,'' in \emph{International Conference on Learning Representations},
  2018, pp. 1--18.

\bibitem{zenke2017continual}
F.~Zenke, B.~Poole, and S.~Ganguli, ``Continual learning through synaptic
  intelligence,'' in \emph{International Conference on Machine Learning}, 2017,
  pp. 3987--3995.

\bibitem{aljundi2018memory}
R.~Aljundi, F.~Babiloni, M.~Elhoseiny, M.~Rohrbach, and T.~Tuytelaars, ``Memory
  aware synapses: Learning what (not) to forget,'' in \emph{European Conference
  on Computer Vision}, 2018, pp. 139--154.

\bibitem{masse2018alleviating}
N.~Y. Masse, G.~D. Grant, and D.~J. Freedman, ``Alleviating catastrophic
  forgetting using context-dependent gating and synaptic stabilization,''
  \emph{Proceedings of the National Academy of Sciences}, vol. 115, no.~44, pp.
  E10\,467--E10\,475, Oct. 2018.

\bibitem{delange2021continual}
M.~Delange, R.~Aljundi, M.~Masana, S.~Parisot, X.~Jia, A.~Leonardis,
  G.~Slabaugh, and T.~Tuytelaars, ``A continual learning survey: Defying
  forgetting in classification tasks,'' \emph{IEEE Transactions on Pattern
  Analysis and Machine Intelligence}, to appear, 2021.

\bibitem{rusu2016progressive}
A.~A. Rusu, N.~C. Rabinowitz, G.~Desjardins, H.~Soyer, J.~Kirkpatrick,
  K.~Kavukcuoglu, R.~Pascanu, and R.~Hadsell, ``Progressive neural networks,''
  \emph{CoRR}, vol. abs/1606.04671, 2016.

\bibitem{fernando2017pathnet}
C.~Fernando, D.~Banarse, C.~Blundell, Y.~Zwols, D.~Ha, A.~A. Rusu, A.~Pritzel,
  and D.~Wierstra, ``{PathNet}: Evolution channels gradient descent in super
  neural networks,'' \emph{CoRR}, vol. abs/1701.08734, 2017.

\bibitem{mallya2018packnet}
A.~Mallya and S.~Lazebnik, ``{PackNet}: Adding multiple tasks to a single
  network by iterative pruning,'' in \emph{IEEE Conference on Computer Vision
  and Pattern Recognition}, 2018, pp. 7765--7773.

\bibitem{mallya2018piggyback}
A.~Mallya, D.~Davis, and S.~Lazebnik, ``Piggyback: Adapting a single network to
  multiple tasks by learning to mask weights,'' in \emph{European Conference on
  Computer Vision}, 2018, pp. 67--82.

\bibitem{rolnick2019experience}
D.~Rolnick, A.~Ahuja, J.~Schwarz, T.~Lillicrap, and G.~Wayne, ``Experience
  replay for continual learning,'' in \emph{Advances in Neural Information
  Processing Systems}, 2019, pp. 350--360.

\bibitem{vitter1985random}
J.~S. Vitter, ``Random sampling with a reservoir,'' \emph{ACM Transactions on
  Mathematical Software}, vol.~11, no.~1, pp. 37--57, Mar. 1985.

\bibitem{lopez2017gradient}
D.~Lopez-Paz and M.~Ranzato, ``Gradient episodic memory for continual
  learning,'' in \emph{Advances in Neural Information Processing Systems},
  2017, pp. 6467--6476.

\bibitem{chaudhry2019efficient}
A.~Chaudhry, M.~Ranzato, M.~Rohrbach, and M.~Elhoseiny, ``Efficient lifelong
  learning with {A-GEM},'' in \emph{International Conference on Learning
  Representations}, 2019, pp. 1--20.

\bibitem{aljundi2019online}
R.~Aljundi, E.~Belilovsky, T.~Tuytelaars, L.~Charlin, M.~Caccia, M.~Lin, and
  L.~Page-Caccia, ``Online continual learning with maximal interfered
  retrieval,'' in \emph{Advances in Neural Information Processing Systems},
  2019, pp. 11\,849--11\,860.

\bibitem{prabhu2020gdumb}
A.~Prabhu, P.~H. Torr, and P.~K. Dokania, ``Gdumb: A simple approach that
  questions our progress in continual learning,'' in \emph{European Conference
  on Computer Vision}, 2020, pp. 524--540.

\bibitem{wu2020subjective}
Q.~Wu, L.~Wang, K.~N. Ngan, H.~Li, F.~Meng, and L.~Xu, ``Subjective and
  objective de-raining quality assessment towards authentic rain image,''
  \emph{IEEE Transactions on Circuits and Systems for Video Technology},
  vol.~30, no.~11, pp. 3883--3897, Nov. 2020.

\bibitem{wang2008maximum}
Z.~Wang and E.~P. Simoncelli, ``Maximum differentiation ({MAD}) competition: A
  methodology for comparing computational models of perceptual quantities,''
  \emph{Journal of Vision}, vol.~8, no.~12, pp. 1--13, Sep. 2008.

\bibitem{choi2015referenceless}
L.~K. Choi, J.~You, and A.~C. Bovik, ``Referenceless prediction of perceptual
  fog density and perceptual image defogging,'' \emph{IEEE Transactions on
  Image Processing}, vol.~24, no.~11, pp. 3888--3901, Nov. 2015.

\bibitem{hayes2020remind}
T.~L. Hayes, K.~Kafle, R.~Shrestha, M.~Acharya, and C.~Kanan, ``Remind your
  neural network to prevent catastrophic forgetting,'' in \emph{European
  Conference on Computer Vision}, 2020, pp. 466--483.

\bibitem{pellegrini2020latent}
L.~Pellegrini, G.~Graffieti, V.~Lomonaco, and D.~Maltoni, ``Latent replay for
  real-time continual learning,'' in \emph{IEEE International Conference on
  Intelligent Robots and Systems}, 2020, pp. 10\,203--10\,209.

\bibitem{hou2019learning}
S.~Hou, X.~Pan, C.~C. Loy, Z.~Wang, and D.~Lin, ``Learning a unified classifier
  incrementally via rebalancing,'' in \emph{IEEE Conference on Computer Vision
  and Pattern Recognition}, 2019, pp. 831--839.

\bibitem{Buzzega2020dark}
P.~Buzzega, M.~Boschini, A.~Porrello, D.~Abati, and S.~Calderara, ``Dark
  experience for general continual learning: A strong, simple baseline,'' in
  \emph{Advances in Neural Information Processing Systems}, vol.~33, 2020, pp.
  15\,920--15\,930.

\bibitem{farquhar2018towards}
S.~Farquhar and Y.~Gal, ``Towards robust evaluations of continual learning,''
  \emph{CoRR}, vol. abs/1805.09733, 2018.

\bibitem{thurstone1927law}
L.~L. Thurstone, ``A law of comparative judgment,'' \emph{Psychological
  Review}, vol.~34, pp. 273--286, Jul. 1927.

\bibitem{tsai2007frank}
M.-F. Tsai, T.-Y. Liu, T.~Qin, H.-H. Chen, and W.-Y. Ma, ``F{R}ank: A ranking
  method with fidelity loss,'' in \emph{International ACM SIGIR Conference on
  Research and Development in Information Retrieval}, 2007, pp. 383--390.

\bibitem{mai2021online}
Z.~Mai, R.~Li, J.~Jeong, D.~Quispe, H.~Kim, and S.~Sanner, ``Online continual
  learning in image classification: An empirical survey,''
  \emph{Neurocomputing}, vol. 469, pp. 28--51, Jan. 2022.

\bibitem{lin2015bilinear}
T.-Y. Lin, A.~RoyChowdhury, and S.~Maji, ``Bilinear {CNN} models for
  fine-grained visual recognition,'' in \emph{IEEE International Conference on
  Computer Vision}, 2015, pp. 1449--1457.

\bibitem{wang2017normface}
F.~Wang, X.~Xiang, J.~Cheng, and A.~L. Yuille, ``{NormFace}: $\mathrm{L}_2$
  hypersphere embedding for face verification,'' in \emph{ACM International
  Conference on Multimedia}, 2017, pp. 1041--1049.

\bibitem{he2016deep}
K.~He, X.~Zhang, S.~Ren, and J.~Sun, ``Deep residual learning for image
  recognition,'' in \emph{IEEE Conference on Computer Vision and Pattern
  Recognition}, 2016, pp. 770--778.

\bibitem{lloyd1982least}
S.~{Lloyd}, ``Least squares quantization in {PCM},'' \emph{IEEE Transactions on
  Information Theory}, vol.~28, no.~2, pp. 129--137, 1982.

\bibitem{masana2020class}
M.~Masana, X.~Liu, B.~Twardowski, M.~Menta, A.~D. Bagdanov, and J.~van~de
  Weijer, ``Class-incremental learning: Survey and performance evaluation on
  image classification,'' \emph{CoRR}, vol. abs/2010.15277, 2020.

\bibitem{sun2016deep}
B.~Sun and K.~Saenko, ``Deep {CORAL}: Correlation alignment for deep domain
  adaptation,'' in \emph{European Conference on Computer Vision}, 2016, pp.
  443--450.

\bibitem{kingma2015adam}
D.~Kingma and J.~Ba, ``Adam: A method for stochastic optimization,'' in
  \emph{International Conference on Learning Representations}, 2015, pp. 1--15.

\bibitem{deng2009imagenet}
J.~Deng, W.~Dong, R.~Socher, L.-J. Li, K.~Li, and L.~Fei-Fei, ``{ImageNet}: A
  large-scale hierarchical image database,'' in \emph{IEEE Conference on
  Computer Vision and Pattern Recognition}, 2009, pp. 248--255.

\bibitem{he2015delving}
K.~He, X.~Zhang, S.~Ren, and J.~Sun, ``Delving deep into rectifiers: Surpassing
  human-level performance on {I}mage{N}et classification,'' in \emph{IEEE
  International Conference on Computer Vision}, 2015, pp. 1026--1034.

\bibitem{ma2017dipiq}
K.~Ma, W.~Liu, T.~Liu, Z.~Wang, and D.~Tao, ``{dipIQ}: Blind image quality
  assessment by learning-to-rank discriminable image pairs,'' \emph{IEEE
  Transactions on Image Processing}, vol.~26, no.~8, pp. 3951--3964, Aug. 2017.

\bibitem{wang2021generalizing}
J.~Wang, C.~Lan, C.~Liu, Y.~Ouyang, and T.~Qin, ``Generalizing to unseen
  domains: {A} survey on domain generalization,'' in \emph{International Joint
  Conference on Artificial Intelligence}, 2021, pp. 4627--4635.

\bibitem{yoon2020scalable}
J.~Yoon, S.~Kim, E.~Yang, and S.~J. Hwang, ``Scalable and order-robust
  continual learning with additive parameter decomposition,'' in
  \emph{International Conference on Learning Representations}, 2020, pp. 1--15.

\bibitem{Bourlard2004AutoassociationBM}
H.~Bourlard and Y.~Kamp, ``Auto-association by multilayer perceptrons and
  singular value decomposition,'' \emph{Biological Cybernetics}, vol.~59, pp.
  291--294, Sept. 2004.

\end{thebibliography}

\end{document}